%% file: main.tex
\documentclass[10pt,twocolumn,letterpaper]{article}

\usepackage[pagenumbers]{cvpr} 

\input{preamble}

\definecolor{cvprblue}{rgb}{0.21,0.49,0.74}
\usepackage[pagebackref,breaklinks,colorlinks,allcolors=cvprblue]{hyperref}
\newcommand\blfootnote[1]{%
    \begingroup
    \renewcommand\thefootnote{}\footnotetext{#1}%
    \endgroup
}

\def\paperID{4}
\def\confName{CVPR}
\def\confYear{2026}

\title{TopoMaskV3: 3D Mask Head with Dense Offset and Height Predictions for Road Topology Understanding}

\author{
Muhammet Esat Kalfaoglu$^{\dagger\S}$ \quad
Halil Ibrahim Ozturk$^{\ddagger}$ \quad
Ozsel Kilinc$^{\S}$ \quad
Alptekin Temizel$^{\dagger}$\\
$^{\dagger}$Graduate School of Informatics, Middle East Technical University, Ankara, Turkey\\
$^{\ddagger}$Togg/Trutek AI Team, Ankara, Turkey
}

\begin{document}
\maketitle
\blfootnote{\raggedright\textbf{Project Page:} \url{https://artest08.github.io/TopoMaskV3.github.io/}.\newline\textbf{Correspondence:} Muhammet Esat Kalfaoglu (esatkalfaoglu@gmail.com, esat.kalfaoglu@metu.edu.tr).\newline\textbf{Contacts:} ibrahim.ozturk@togg.com.tr; ozsel@amazon.co.uk; atemizel@metu.edu.tr.\newline $^{\S}$ Muhammet Esat Kalfaoglu and Ozsel Kilinc conducted this work at Togg/Trutek AI and are currently with Ultralytics and Amazon, respectively.}

\input{topomaskv3/abstract}
\input{topomaskv3/introduction}
\input{topomaskv3/related_work}

\input{topomaskv3/methodology}
\input{topomaskv3/experiments}
\input{topomaskv3/conclusion}

\clearpage
{
    \small
    \bibliographystyle{ieeenat_fullname}
    \bibliography{main}
}

\clearpage
\input{topomaskv3/supplementary}

\end{document}

%% file: preamble.tex
\usepackage{graphicx}
\usepackage{amsmath}
\usepackage{amssymb}
\usepackage{booktabs}
\usepackage{bbm}
\usepackage{multirow}
\usepackage{enumitem}
\setitemize{noitemsep,topsep=0pt,parsep=0pt,partopsep=0pt, leftmargin=*}
\DeclareMathAlphabet\mathbfcal{OMS}{cmsy}{b}{n}

\newcommand{\beginsupplement}{
    \setcounter{table}{0}
    \renewcommand{\thetable}{S\arabic{table}}
    \setcounter{figure}{0}
    \renewcommand{\thefigure}{S\arabic{figure}}
    \setcounter{equation}{0}
    \renewcommand{\theequation}{S\arabic{equation}}
}

%% file: topomaskv3/abstract.tex
\begin{abstract}
Mask-based paradigms for road topology understanding, such as TopoMaskV2, offer a
complementary alternative to query-based methods by generating centerlines via a
dense rasterized intermediate representation. However, prior work was limited to 2D
predictions and suffered from severe discretization artifacts, necessitating fusion
with parametric heads. We introduce \textbf{TopoMaskV3}, which advances this pipeline
into a robust, standalone 3D predictor via two novel dense prediction heads: a
\textbf{dense offset field} for sub-grid discretization correction within the existing
BEV resolution, and a \textbf{dense height map} for direct 3D estimation. Beyond the
architecture, we are the \textbf{first} to address geographic data leakage in road
topology evaluation by introducing (1) \textbf{geographically distinct splits} to prevent
memorization and ensure fair generalization, and (2) a \textbf{long-range} ($\pm$100\,m)
benchmark. \textbf{TopoMaskV3 achieves state-of-the-art 28.5~OLS} on this geographically
disjoint benchmark, surpassing all prior methods. Our analysis shows that the mask
representation is more robust to geographic overfitting than Bezier, while LiDAR fusion is
most beneficial at long range and exhibits larger relative gains on the overlapping original
split, suggesting overlap-induced memorization effects.
\end{abstract}

%% file: topomaskv3/introduction.tex
\section{Introduction}
\label{sec: introduction}

Road topology understanding transcends the mere detection of static road elements like lanes
or traffic lights, fundamentally requiring \textbf{reasoning about their complex
inter-relationships}. Pure detection is insufficient for robust autonomous navigation;
a model must infer connectivity (such as how lanes merge or divide), determine valid paths
through complex road intersections, and correctly associate traffic control elements with
the specific lanes they govern.

TopoMaskV2~\cite{kalfaoglu_topomaskv2_2024} introduced a novel paradigm for road topology by
generating centerlines directly from a rasterized mask. Its core innovation was enriching
the mask representation with a quad-direction label, providing the crucial flow information
necessary to convert a mask representation into an ordered, directed polyline. While
inspiring, the mask-based pipeline had significant limitations: it suffered from
\textbf{discretization artifacts} during raster-to-vector conversion and, lacked
\textbf{height prediction}, restricting it to 2D output. To compensate for these
limitations and achieve competitive performance, TopoMaskV2 ultimately relied on fusing its
modest mask-based results with a parallel, and more robust parametric (Bezier) head.

This paper introduces TopoMaskV3, which directly addresses these core limitations and
advances the mask-based pipeline into a robust, \textbf{standalone 3D predictor}. The
architecture has two novel components: a \textbf{dense offset field}
(Section~\ref{sec: proposal_mechanisms}) to correct discretization errors and achieve
sub-grid accuracy, and a \textbf{dense height map} (Section~\ref{sec: height_prediction})
to predict the $z$-coordinate. As a result, this shifts the mask head from a weak
module that required fusion-based compensation to a highly competitive, standalone 3D
perception method.

Existing benchmarks for road topology understanding rely on temporal splits, which are
suitable for dynamic objects but fundamentally problematic for static elements. Since the
underlying map is invariant, data collected at different times revisit the same geographic
locations, resulting in \textbf{geographic overlap} between training and testing data
\cite{yuan_streammapnet_2024, lilja_localization_2024}. Building on the geographically
distinct splitting methodology proposed by \cite{yuan_streammapnet_2024,
lilja_localization_2024} for HD map tasks, we are the \textbf{first to adapt this critical
evaluation framework to road topology understanding}. This adaptation is substantially
harder: since the official OpenLane-V2 test set is not publicly available, we replicated the
entire ground-truth generation pipeline from raw Argoverse~2 HDMap data, handling the
topological complexity of interconnected directed graphs. This rigorous effort yields a
stable measure of true generalization for road topology models.
In addition to geographically fair splits, we also introduce, for the first time, a
\textbf{long-range benchmark} that extends the conventional $\pm 50\text{ m}$ range to a
more challenging $\pm 100\text{ m}$. This extension provides a more rigorous benchmark to
analyze model robustness and generalization, since error propagation and performance trends
can change markedly and become severely amplified at extended distances. This setting is
especially useful for measuring the real benefit of LiDAR fusion, and is also relevant for
future SD map and satellite-image studies.

Under this combined benchmark setting (non-overlapping splits and the extended
$\pm 100\text{ m}$ range), we reassess prior architectural assumptions about
\textbf{output fusion} (Mask vs. Bezier) and \textbf{sensor fusion} (Camera vs. LiDAR). This
is particularly important because LiDAR-based models can overfit under overlap-prone
original splits. Our experiments (Section~\ref{sec: sensor_modalities_long_range} and
Section~\ref{sec: output_types_dataset_splits}) resolve the ambiguity around the utility of
output fusion \cite{kalfaoglu_topomaskv2_2024}: the fused output provides a modest advantage
on realistic, geographically distinct splits. More critically, we demonstrate that sensor
fusion delivers a substantially larger improvement in the more challenging long-range
setting \cite{kalfaoglu_topobda_2024}. Together, these results provide the key insight that
both fusion types are essential mechanisms for achieving robust generalization and better
performance, particularly at extended distances. The main contributions of this work are:


\begin{itemize}[nosep]
\item \textbf{A Standalone 3D Mask-Based Predictor:} We extend the mask-based paradigm by
introducing two novel, dense prediction heads: \textbf{a dense offset field} to accurately
correct inherent discretization artifacts, and a \textbf{dense height map} to enable robust,
end-to-end 3D centerline prediction, transforming the mask head into a highly competitive,
standalone method.
\item \textbf{A Rigorous Generalization Benchmark for Road Topology:} Building on geographic
splitting concepts from HD map benchmarking~\cite{yuan_streammapnet_2024,
lilja_localization_2024}, we are the \textbf{first} to address geographic data leakage in
road topology evaluation. This rigorous evaluation transitions assessment from rewarding
\textbf{geographic memorization} to measuring true structural generalization.
\item \textbf{The Long-Range Challenge:} We introduce, for the first time, a long-range
benchmark that significantly extends the evaluation scope from the conventional
$\pm 50\text{ m}$ to a challenging $\pm 100\text{ m}$, providing a crucial measure of model
robustness at extended distances required for high-speed driving.
\item \textbf{Key Insights on Fusion:} We conduct a comprehensive analysis of output fusion
(Mask vs. Bezier) and sensor fusion (Camera vs. LiDAR) on these new, demanding benchmarks,
demonstrating that both fusion types are essential for achieving robust generalization and
better performance, particularly in the long-range setting.
\end{itemize}

%% file: topomaskv3/related_work.tex
\section{Related Work}
\label{sec: related_work}

\subsection{Road Topology Understanding}
The fundamental challenge in road topology understanding is to detect static road elements while simultaneously inferring the relational structure and connectivity between them. STSU~\cite{can_structured_2021} introduced a method to extract a directed road graph from a single image. TopoNet~\cite{li_graph-based_2023} was a seminal work that established a strong benchmark and proposed a Graph Neural Network (GNN) based framework to explicitly model relational knowledge. LaneSegNet~\cite{li_lanesegnet_2024} expanded the task scope, introducing the ``lane segment" concept to jointly predict not only centerlines but also lane dividers and drivable areas.

A dominant paradigm subsequently emerged using end-to-end transformers to predict vectorized representations, grouped by their output format:
\begin{itemize}[nosep]
\item\textbf{Parametric Representations using Bezier curves:} This strategy models road elements using compact, mathematically defined curves. TopoMLP~\cite{wu_topomlp_2024} was foundational in demonstrating the effectiveness of the parametric Bezier curve representation instead of keypoint prediction \cite{li_graph-based_2023, li_lanesegnet_2024} for predicting centerlines.

\item\textbf{Path-Wise Representations:} A key conceptual advance came from LaneGAP~\cite{liao_lane_2023}, which proposed predicting holistic, continuous paths that span intersections, thereby better preserving lane continuity, building upon the point-query-based architecture of MapTR~\cite{liao_maptr_2023}. This path-wise modeling was later improved by MapTRV2~\cite{liao_maptrv2_2024}.

\item\textbf{Mask-Based Raster-to-Vector:} TopoMaskV2~\cite{kalfaoglu_topomaskv2_2024} introduced a mask-based paradigm, generating directed centerlines from a rasterized mask via a quad-direction label. Crucially, this approach was limited by discretization artifacts and its 2D-only output, necessitating fusion with Bezier heads.

\item\textbf{Specialized Reasoning and Feature Enhancements:}  Topo2D~\cite{li_enhancing_2024} focuses on feature enhancement by fusing 2D lane priors to aid 3D learning. Other works improve the reasoning process itself: TopoFormer~\cite{lv_t2sg_2024} uses geometric-aware attention to enhance relational reasoning. Addressing the critical "endpoint deviation" problem, TopoLogic~\cite{fu_topologic_2024} introduced an interpretable pipeline that combines geometric distances and semantic similarity. TopoPoint~\cite{fu_topopoint_2025} proposes to solve this by explicitly detecting the endpoints as independent queries.

\end{itemize}

\subsection{Multi-modal and Temporal Road Topology Understanding}
\begin{itemize}[nosep]
\item\textbf{Multi-Modality:} A significant trend is the use of Standard Definition (SD) map priors. SMERF~\cite{luo_augmenting_2023} tokenizes SD map elements for a transformer decoder, TopoSD~\cite{yang_toposd_2024} enriches BEV features with map tokens, and SMART~\cite{ye_smart_2025} uses both SD and satellite maps. TopoBDA~\cite{kalfaoglu_topobda_2024} combines SD map features with LiDAR sensor, demonstrating the benefits of multi-modal fusion. 
\item\textbf{Temporal Context:} This line of work aims to address temporal stability. TopoStreamer~\cite{yang_topostreamer_2025} maintains a ``topology-aware state" for consistent connectivity across frames, while FASTopoWM~\cite{yang_fastopowm_2025} employs a fast-slow latent world model to provide stable temporal context.
\end{itemize}


\subsection{Benchmarking for Generalization}
The problem of geographic data leakage in static map benchmarks was recently highlighted by StreamMapNet~\cite{yuan_streammapnet_2024}, which showed that temporal splits cause \textbf{geographic overlap} and reward memorization. Building on this, \cite{lilja_localization_2024} proposed a formal set of \textbf{geographically distinct splits} to rigorously test generalization. While these foundational works focused on HD map element prediction, our study is the first to adapt this crucial splitting methodology to the road topology understanding benchmarks.

%% file: topomaskv3/methodology.tex
\section{Methodology}
\label{sec: methodology}


\begin{figure}[tb]
  \centering
  \includegraphics[width=0.75\linewidth]{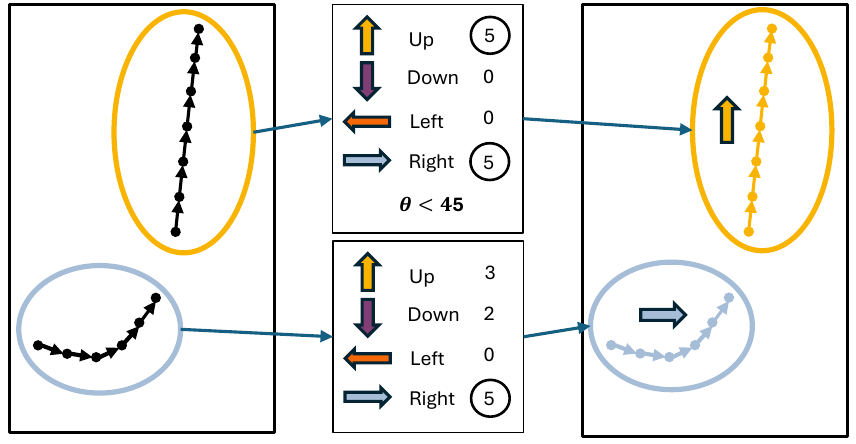}
  \caption{\textbf{Quad-Direction Labels Encoding.} Each centerline is assigned one of four directional labels: up, down, left, or right, based on majority voting between consecutive points. Ties are resolved using the angle between the start and end points.}
  \label{fig: quad_direction_label_representation_recap}
\end{figure}


\begin{figure*}[t]
  \centering
  \includegraphics[width=0.90\textwidth]{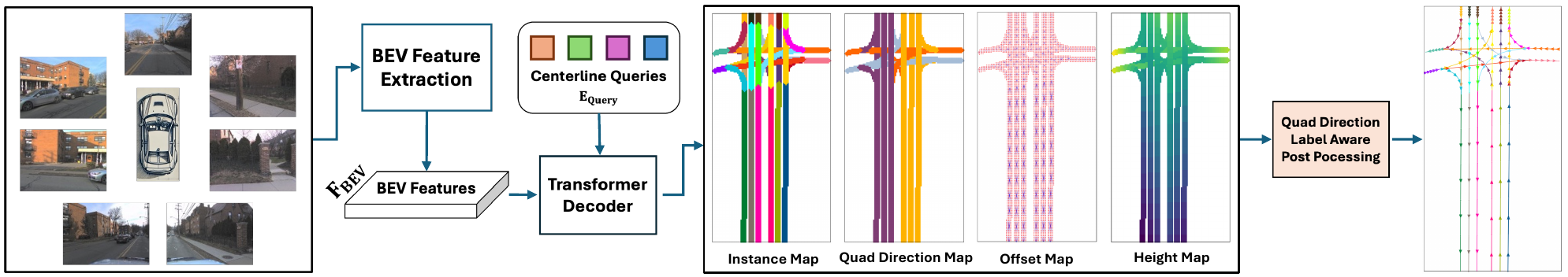}
  \caption{\textbf{TopoMaskV3 Architecture Overview.} The method adopts an instance-query-based design. Bird's Eye View (BEV) features extracted from multi-camera images are processed by a transformer decoder that predicts: binary masks, quad-direction labels, 2D offsets, and height maps. A quad-direction-aware post-processing step then converts these dense outputs into \textbf{3D centerline instances}.}
  \label{fig: topomaskv3_overview}
  \vspace{-0.5cm}
\end{figure*}
To incorporate flow information into road topology masks, we adopt the quad-direction label representation \cite{kalfaoglu_topomaskv2_2024}. Each centerline instance is assigned a single direction label (\textit{up, down, left, right}) through a voting mechanism across consecutive points (Fig.~\ref{fig: quad_direction_label_representation_recap}). These labels serve as semantic cues for flow-aware mask instances and are trained using a cross-entropy loss. 

The overview of the TopoMaskV3 method is shown in Fig.~\ref{fig: topomaskv3_overview}. TopoMaskV3 constructs a unified Bird’s Eye View (BEV) representation from multi-camera RGB inputs. The unified BEV features are processed by a transformer decoder equipped with sparse queries, each corresponding to a candidate centerline instance. For each query, the model predicts a set of outputs including quad-direction labels, instance masks, height maps, and offset fields. These outputs are then post-processed to refine and generate 3D centerline points with associated flow information.




\subsection{Multi-View to BEV Projection}
The pipeline begins by processing the perspective-view images $\{\mathbf{I}_i\}_{i=1}^N$, $\mathbf{I}_i \in \mathbb{R}^{H_I \times W_I \times 3}$ to generate a shared BEV feature map $\mathbf{F}_{BEV}$. First, a shared backbone encoder $f_{PV}$ extracts perspective-view features $\mathbf{F}_{PV_i} = f_{PV}(\mathbf{I}_i)$, $\mathbf{F}_{PV_i} \in \mathbb{R}^{H_{PV} \times W_{PV} \times C_{PV}}$. These features are then aggregated and projected into the top-down BEV feature map via a projection module $f_{BEV}$, yielding $\mathbf{F}_{BEV} = f_{BEV}(\{\mathbf{F}_{PV_i}\}_{i=1}^N)$, $\mathbf{F}_{BEV} \in \mathbb{R}^{H_{BEV} \times W_{BEV} \times C_{BEV}}$. The projection function $f_{BEV}$ can be instantiated using Lift-Splat-Shoot~\cite{philion_lift_2020, huang_bevdet_2021, li_bevdepth_2023}, transformer-based designs~\cite{li_bevformer_2024, chen_efficient_2022, zhou_cross-view_2022, wang_exploring_2023}, or other efficient BEV encoders~\cite{li_fast-bev_2024, xie_m2bev_2022, harley_simple-bev_2023}. This step consolidates multi-view spatial cues into a compact top-down representation suitable for downstream processing.

\subsection{Prediction Heads of TopoMaskV3}
\label{sec: prediction_heads_of_topomaskv3}

\begin{figure}[tb]
  \centering
  \includegraphics[width=0.99\linewidth]{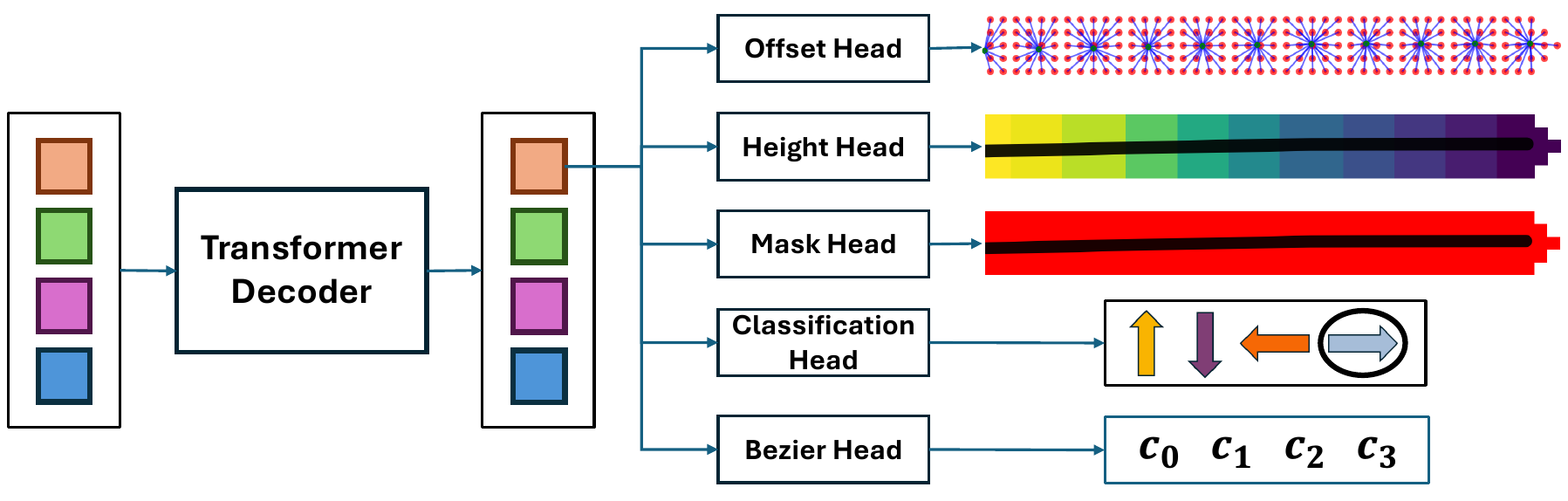}
  \caption{\textbf{TopoMaskV3 Decoder Architecture.} Each sparse query is decoded by five parallel heads, each predicting a different centerline attribute.}
  \label{fig: topomask_v3_decoder_heads}
\end{figure}

TopoMaskV3 utilizes a sparse query design, where each query corresponds to a distinct centerline instance. As illustrated in Figure~\ref{fig: topomask_v3_decoder_heads}, the decoder has five prediction heads:
\begin{itemize}[nosep]
\item \textbf{Classification Head:} Predicts the quad-direction label, which is critical for defining the directional flow, ordering the refined points, and determining the primary axis of polynomial fitting (Section \ref{sec: curve_reconstruction}). 
\item \textbf{Mask Head:} Produces the mask probability map $\mathbf{M}_{prob}$. This map is binarized to generate the rasterized mask $\mathbf{R}$ for centerline extraction.
\item \textbf{Offset Head:} Predicts the dense 2D offset field $\mathbf{O} \in \mathbb{R}^{H_{BEV} \times W_{BEV} \times 2}$, which is used to achieve sub-grid accuracy by correcting discretization artifacts (Section~\ref{sec: proposal_mechanisms}).
\item \textbf{Height Head:}  Estimates the dense height map $\mathbf{H} \in \mathbb{R}^{H_{BEV} \times W_{BEV}}$. This map provides the $z$-coordinate for 3D prediction, sampled at the final $(x,y)$ location of each refined centerpoint.
\item \textbf{Bezier Head:}  Outputs a set of 3D Bezier control points, providing an alternative, parametric representation of the centerline.
\end{itemize}

The architecture provides two distinct paths for generating the final 3D centerline:
\begin{itemize}[nosep]
\item 
\textbf{Primary mask-based path} synthesizes the outputs from the \emph{Classification}, \emph{Mask}, \emph{Offset}, and \emph{Height} heads. These outputs are processed through the full \emph{Curve Reconstruction} pipeline (Section~\ref{sec: curve_reconstruction}) to yield a final set of points, $\mathcal{P}_{mask}$.
\item 
\textbf{Bezier-based path (optional)} generates a parametric curve directly by sampling $N$ ordered 3D points, $\mathcal{P}_{bezier}$, from the output of the \emph{Bezier Head}.
\end{itemize}

The Bezier head is not required for the baseline model (which uses the primary path), but it is essential for two specific extensions: (i) replacing the baseline Masked Attention (MA) with Bezier Deformable Attention (BDA)~\cite{kalfaoglu_topobda_2024} (See Section~\ref{sup_sec: masked_and_bezier_deformable_attention_recap}), or (ii) enabling the output fusion mechanism described below.

When the Bezier head is activated, an optional output fusion step can be benefited from to leverage the complementary strengths of the mask-based and parametric representations. Let $\mathcal{P}_{mask} = \{\mathbf{m}_i\}_{i=1}^N$ and $\mathcal{P}_{bezier} = \{\mathbf{b}_i\}_{i=1}^N$ be the set of $N$ ordered 3D points from the mask-based and the Bezier-based path, respectively, where $\mathbf{m}_i = (x_i^m, y_i^m, z_i^m)$.
and $\mathbf{b}_i = (x_i^b, y_i^b, z_i^b)$. The  coordinates of these two corresponding point sets $\mathcal{P}_{mask}$ and $\mathcal{P}_{bezier}$ are averaged element-wise to obtain the fused point $\mathbf{f}_i$ (Eq. \ref{eq: fusion}).

\begin{equation}
    \mathbf{f}_i = \left( \frac{x_i^m + x_i^b}{2}, \frac{y_i^m + y_i^b}{2}, \frac{z_i^m + z_i^b}{2} \right)
\label{eq: fusion}
\end{equation}

\subsection{Proposal Mechanisms for Offset Refinement}
\label{sec: proposal_mechanisms}

At inference, the decoder predicts a probability map $\mathbf{M}_{prob}\in[0,1]^{H_{BEV}\times W_{BEV}}$ from the BEV feature map $\mathbf{F}_{BEV}$. This map is binarized to obtain a rasterized mask $\mathbf{R}$ via thresholding (Eq. \ref{eq: rasterization}). 
\begin{equation}
    \mathbf{R}(i,j) = \mathbbm{1}\big[\mathbf{M}_{prob}(i,j) \ge \tau\big], \quad \tau\in(0,1)
    \label{eq: rasterization}
\end{equation}
where $i$ and $j$ index the grid rows and columns, respectively. Then, a set of coarse centerpoints is extracted from $\mathbf{R}$ based on the quad-direction label:



The extraction strategy depends on the assigned direction. For `up'/`down', \emph{row-wise expectation} is applied over all columns $j$ (Eq.~\ref{eq: row_expectation}) to obtain point $\hat{\mathbf{p}}(i) = (i,\hat{j}(i))$; for `left'/`right', \emph{column-wise expectation} is applied over all rows $i$ (Eq.~\ref{eq: col_expectation}) to obtain point $\hat{\mathbf{p}}(j) = (\hat{i}(j), j)$.

\begin{subequations} 
\noindent\begin{minipage}{.49\linewidth} 
    \begin{equation}
        \hat{j}(i) =
        \frac{\sum\limits_{j=0}^{W_{BEV}-1} \mathbf{R}(i,j)\cdot j}
             {\sum\limits_{j=0}^{W_{BEV}-1} \mathbf{R}(i,j)}
        \label{eq: row_expectation}
    \end{equation}
\end{minipage}
\hfill 
\begin{minipage}{.49\linewidth} 
    \begin{equation}
        \hat{i}(j) =
        \frac{\sum\limits_{i=0}^{H_{BEV}-1} \mathbf{R}(i,j)\cdot i}
             {\sum\limits_{i=0}^{H_{BEV}-1} \mathbf{R}(i,j)}
        \label{eq: col_expectation}
    \end{equation}
\end{minipage}
\end{subequations}

\begin{figure}[!t]
  \centering
  \includegraphics[width=0.70\linewidth]{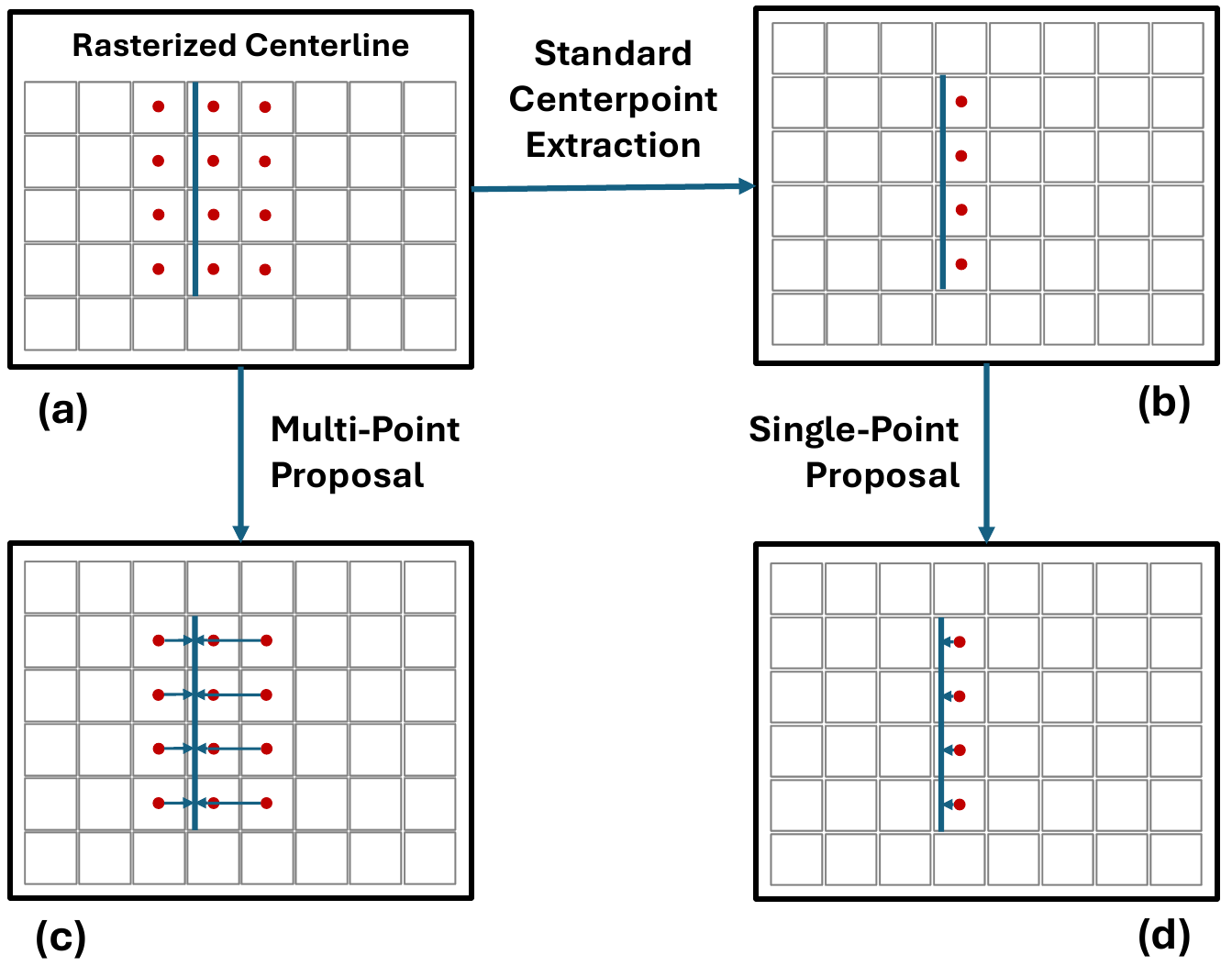}
  \caption{\textbf{Offset Refinement Scheme.}
          (a) A continuous straight centerline (blue) and its rasterized representation.
          (b) Centerpoints obtained using conventional row/column-wise extraction
          (c) Multi-point proposal predicts an offset vector for each raster pixel to its closest point on the continuous centerline, enabling one-to-many matching
          (d) Single-point proposal refines the centerpoints by predicting offsets toward their nearest centerline point, enforcing one-to-one matches, and refining centerline localization.
    }
  \label{fig: offset_refinement}
\end{figure}
 
This baseline extraction process of continuous centerlines from rasterized masks (Fig.~\ref{fig: offset_refinement}a to Fig.~\ref{fig: offset_refinement}b) is susceptible to discretization artifacts (or gridization error), as the true centerline rarely aligns perfectly with grid cell centers (Fig.~\ref{fig: offset_refinement}b) 

To achieve sub-grid accuracy, we propose to learn a dense \textbf{offset field} $\mathbf{O}\in\mathbb{R}^{H_{BEV}\times W_{BEV}\times 2}$, where each grid cell $(i,j)$ predicts a 2D offset vector $\mathbf{o}_{ij} = \mathbf{O}(i,j,:)$.

The offset field is trained using a \textbf{multi-point supervision} strategy (Fig.~\ref{fig: offset_refinement}c). For every foreground pixel $(i,j)$ in the \emph{ground-truth} rasterized mask, the network learns to predict an offset $\mathbf{o}_{ij}$ that points from the grid center $(i,j)$ to the closest point on the \emph{continuous ground-truth centerline} $\mathcal{C}$. The target offset $\mathbf{o}_{ij}^{gt}$ is defined in Eq. \ref{eq:TargetOffset}.
\begin{equation}
\label{eq:TargetOffset}
\mathbf{o}_{ij}^{gt} = \Pi_{\mathcal{C}}((i,j)) - (i,j), \quad \text{for all } \mathbf{R}(i,j)=1    
\end{equation}

where $\Pi_{\mathcal{C}}(\mathbf{x}) = \arg\min_{\mathbf{c}\in\mathcal{C}}\|\mathbf{x}-\mathbf{c}\|_2$ is the projection operator finding the closest point on the continuous curve $\mathcal{C}$. This dense, one-to-many matching scheme ensures the network learns to correct discretization error from any point on the mask. At inference, we utilize the predicted offset map $\mathbf{O}$ and the predicted rasterized map $\mathbf{R}$ (Eq.~\eqref{eq: rasterization}) in two alternative ways:

\begin{itemize}[nosep]
    \item \textbf{Single-Point Proposal:} This approach refines the \emph{baseline} points. First, we extract the coarse centerpoints $\hat{\mathbf{p}}_k = ({i}_k,{j}_k)$ using the direction-aware expectations (Eqs.~\eqref{eq: row_expectation}–\eqref{eq: col_expectation}). Then, we retrieve the predicted offset $\mathbf{o}_k = \mathbf{O}({i}_k,{j}_k,:)$ at each point's location and apply the refinement: $\tilde{\mathbf{p}}_k = \hat{\mathbf{p}}_k + \mathbf{o}_k.$
    This \textbf{one-to-one} scheme (Fig.~\ref{fig: offset_refinement}d) efficiently corrects only the initially extracted points.

    \item \textbf{Multi-Point Proposal:} This approach applies the training logic directly at inference. \emph{All} foreground pixels in the predicted mask $\mathbf{R}$ are refined using their corresponding offsets (Eq. \ref{eq:Refine}).
    \begin{equation}
        \tilde{\mathbf{p}}_{ij} = (i,j) + \mathbf{o}_{ij}, \quad \text{for all } \mathbf{R}(i,j)=1
        \label{eq:Refine}
    \end{equation}
    This \textbf{one-to-many} scheme (Fig.~\ref{fig: offset_refinement}c) provides a dense correction across the entire mask region.
\end{itemize}


\subsection{Height Prediction}
\label{sec: height_prediction}
To elevate the 2D centerlines into 3D, the network predicts the dense \textbf{height map} $\mathbf{H} \in \mathbb{R}^{H_{BEV} \times W_{BEV}}$ which assigns a normalized height value $h = \mathbf{H}(i,j)$ to each grid cell.

\vspace{0.3em}
\noindent\textbf{Height Map Supervision.}
The training for the height map follows the same \textbf{multi-point, closest-point} principle used for offset supervision. 

For every foreground pixel in the \emph{ground-truth} rasterized mask, the network is trained to predict the normalized height $h_{norm}(\mathbf{c})$ of the closest point $\mathbf{c}$ on the continuous 3D centerline $\mathcal{C}$. The target height $h_{ij}^{gt}$ is defined as in Eq. \ref{eq: TargetHeight}.
\begin{equation}
h_{ij}^{gt} = h_{norm}\left( \Pi_{\mathcal{C}}((i,j)) \right), \quad \text{for all } \mathbf{R}(i,j)=1  
\label{eq: TargetHeight}    
\end{equation}

where $\Pi_{\mathcal{C}}(\mathbf{x}) = \arg\min_{\mathbf{c}\in\mathcal{C}}\|\mathbf{x}-\mathbf{c}\|_2$ is the same projection operator used for offset supervision (Section~\ref{sec: proposal_mechanisms}). This ensures that all pixels in the rasterized region learn to predict the correct height of the nearest true centerline point.

\vspace{0.3em}
\noindent\textbf{Height Assignment at Inference.}
At inference, the predicted height map $\mathbf{H}$ is used to provide the normalized height value $h$ for the points generated by the two alternative proposal mechanisms. This value is sampled from the point's original grid location, in the same way the offset vector is retrieved.

\textbf{Single-Point Height Assignment:} For the \emph{single-point proposal}, we use the coarse centerpoint locations $\hat{\mathbf{p}}_k = ({i}_k,{j}_k)$ derived from the expectation step (Eqs.~\eqref{eq: row_expectation}–\eqref{eq: col_expectation}). The height for each point $k$ is sampled directly from the map $h_k = \mathbf{H}({i}_k,{j}_k)$

This height value $h_k$ is appended to its corresponding 2D offset-refined location $(\tilde{p}_{k,i}, \tilde{p}_{k,j})$ (from Section~\ref{sec: proposal_mechanisms}) to form the final 3D point $\tilde{\mathbf{p}}_k = (\tilde{p}_{k,i}, \tilde{p}_{k,j}, h_k)$.

\textbf{Multi-Point Height Assignment:} For the \emph{multi-point proposal}, all foreground pixels in the predicted mask region $\mathbf{R}$ are processed. Each 3D grid point $\tilde{\mathbf{p}}_{ij}$ is constructed as in Eq. \ref{eq: 3DGridPoint}.
\begin{equation}
        \tilde{\mathbf{p}}_{ij} = (i+o_{ij,x}, j+o_{ij,y}, h_{ij}), \quad \text{for all } \mathbf{R}(i,j)=1  
\label{eq: 3DGridPoint}
\end{equation}
    where $\mathbf{o}_{ij} = (o_{ij,x}, o_{ij,y})$ is the 2D offset vector from $\mathbf{O}(i,j)$ and $h_{ij} = \mathbf{H}(i,j)$ is the height value.

In both cases, this step generates a set of 3D points (either $\{\tilde{\mathbf{p}}_k\}$ or $\{\tilde{\mathbf{p}}_{ij}\}$), which we denote generically as $\{\tilde{\mathbf{p}}_{grid}\}$. This set serves as the input to the Curve Reconstruction pipeline described in Section~\ref{sec: curve_reconstruction}.

\subsection{Curve Reconstruction}
\label{sec: curve_reconstruction}
The final set of refined 3D grid points $\{\tilde{\mathbf{p}}_{grid}\}$ generated by the offset and height heads must be transformed and regularized into a smooth, ordered, continuous curve in real-world coordinates. This process involves two main stages:
\noindent\textbf{1. Coordinate Transformation:} First, each refined grid point $\tilde{\mathbf{p}}_{grid} = (i, j, h)$ is mapped to its real-world 3D coordinate $\tilde{\mathbf{p}}_{world} = (x, y, z)$ via a pre-defined grid-to-world transformation matrix $\mathbf{V}^{-1}$, yielding a set of unordered 3D points.

\noindent\textbf{2. Direction-Aware Regularization and Ordering:} Starting from the noisy real-world points ${\tilde{\mathbf{p}}_{world}}$, we produce a smooth, ordered 3D centerline through three steps.
\textbf{Polynomial Fitting.} Separate polynomial functions are fit to the noisy points. The \textbf{quad-direction label} dictates the independent variable for the 2D path fit (e.g., $y=f(x)$ for `up'/`down' directions), simultaneously, a 3D height surface ($z=g(x,y)$) is also fit.
\textbf{Resampling.} The resulting 2D path is sampled and then resampled using \textbf{arc-length interpolation} to generate a final set of $N$ equidistant 3D points.
\textbf{Ordering.} This final set of $N$ points is then \textbf{explicitly sorted} (e.g., increasing $x$ for `up') to ensure the correct directional flow.

This joint regularization and interpolation pipeline suppresses the discretization artifacts further and ensures a high-quality, continuous vector output. Implementation details of this pipeline are given in Section~\ref{sup_sec: curve_reconstruction_details}.

\subsection{LiDAR Sensor Fusion}
LiDAR data is incorporated using a voxel-space fusion pipeline \cite{kalfaoglu_topobda_2024}. First, perspective-view camera features are projected into a voxel grid. Concurrently, the raw LiDAR point cloud is processed by a dedicated LiDAR encoder to produce a separate, dense LiDAR voxel grid. These two voxel representations are then concatenated along the channel dimension before being collapsed into a final, unified BEV feature map. A detailed mathematical formulation of this pipeline is provided in Section~\ref{sup_sec: lidar_fusion_details}.

%% file: topomaskv3/experiments.tex
\section{Experimental Evaluations}
\label{sec: experimental_evaluation}

\noindent\textbf{Dataset:}
The original OpenLane-V2 dataset \cite{wang_openlane-v2_2024}, which is derived from the Argoverse2 \cite{wilson_argoverse_2023} (Subset-A) and NuScenes \cite{caesar_nuscenes_2020} (Subset-B) datasets, has a significant shortcoming: its standard splits were designed for dynamic object detection and \textbf{do not prevent geographic overlap} between training and validation sets, a flaw highlighted by recent studies \cite{yuan_streammapnet_2024, lilja_localization_2024}.
To enable rigorous evaluation, we first developed a custom annotation pipeline using the HDMap data from Argoverse2 to replicate the OpenLane-V2 ground truth structure. Critically, and building on the concept of \textbf{geographically disjoint splits} \cite{lilja_localization_2024}, we then introduce five new and distinct splits for Subset-A specifically tailored for road topology understanding: \textbf{Original}, \textbf{Near}, \textbf{FarA}, \textbf{FarB}, and \textbf{FarC}. These splits are designed to rigorously test model generalization and overcome the data leakage inherent in the original benchmark. Details of these evaluation protocols are provided in Table~\ref{tab: dataset_splits}. Release details are summarized in Supplementary Section~\ref{sup_sec: benchmark_release}.

\begin{table}[t]
    \centering
    \caption{Proposed Dataset Splits for Road Topology Evaluation.}
    \label{tab: dataset_splits}
    \scalebox{0.65}{
        \begin{tabular}{lccc}
            \toprule
            \textbf{Dataset Split} & \textbf{Geo. Overlap} & \textbf{Shared Cities} & \textbf{Evaluation Focus} \\
            \midrule
            \textbf{Original} & Yes & Yes & In-distribution performance \\
            \textbf{Near} & No & Yes & Within-city generalization  \\
            \textbf{Far A, B, C} & No & No & Robust Geographic Generalization\\
            \bottomrule
        \end{tabular}
    }
\end{table}


\noindent\textbf{Metric:}
For centerline detection, two distinct evaluation metrics are employed. The $\text{DET}_{l}$ metric, based on the Fréchet distance, captures both the spatial proximity and directional alignment between the predicted and ground-truth centerlines. $\text{DET}_{l\_ch}$ metric, based on Chamfer distance, evaluates only the spatial closeness, disregarding point ordering and directionality. Both metrics are computed using a Mean Average Precision (mAP) formulation over a set of thresholds.

The $\text{TOP}_{ll}$ metric, an Average Precision (AP)-based score for topology evaluation, suffers from a critical flow in its standard V1.1 implementation: it uses a fixed confidence threshold of $0.5$ for ranking, which non-standardly truncates the precision-recall curve. A detailed analysis of this thresholding flaw is provided in Section~\ref{sup_sec: topology_metric}. 


This flaw can complicate internal comparisons, as demonstrated in~\cite{kalfaoglu_topomaskv2_2024}. Therefore, the score remapping technique ($P(x) \rightarrow P(x) + 1 \times [P(x) > 0.05]$) is adopted to ensure a stable comparison \emph{within our ablation studies} without modifying the evaluation pipeline. However, for the main SOTA comparison in Table~\ref{tab: sota_camera_only_subsetA}, the original, unmodified TOP\textsubscript{ll} scores are reported to maintain fair comparability with prior literature.


A single, comprehensive score for centerline-focused road topology is provided by the $\text{OLS}_{l}$ metric (Eq.~\ref{Eq: OLS_l}), as in~\cite{kalfaoglu_topobda_2024}.
\begin{equation}
\label{Eq: OLS_l}
    \text{OLS}_{l} = \frac{1}{3} \bigg[ \text{DET}_{l} + \text{DET}_{l\_ch} + f(\text{TOP}_{ll}) \bigg]
\end{equation}
This metric is adapted from the original OpenLaneV2 Score (OLS)~\cite{li_graph-based_2023} by omitting traffic element-related components ($\text{DET}_{t}$ and $\text{TOP}_{lt}$) and incorporating the Chamfer distance component ($\text{DET}_{l\_ch}$). This adaptation is necessary for two key reasons: it provides a more focused assessment of centerline geometry and topology, and the ground-truth annotations for traffic elements are not available for some samples in the new geographically distinct splits.

\noindent\textbf{Implementation Details:}
For the mask-based pipeline, the probability threshold $\tau$ (Eq.~\ref{eq: rasterization}) is $0.95$. For curve reconstruction (Section~\ref{sec: curve_reconstruction}), a 4th-order polynomial with arc-length interpolation is used. These hyperparameters are justified by ablations in Sections~\ref{sup_sec: post_process_impact} and \ref{sup_sec: threshold_value_impact_on_results}. Full details on the model architecture, optimization, and preprocessing are available in Section~\ref{sup_sec: implementation_details}, and the complete loss function definitions are provided in Section~\ref{sup_sec: experiments_loss_functions}.

\subsection{Comparison of Offset Refinement Strategies}
\label{sec: comparison_of_proposal_mechanism_for_offset_refinements}
\begin{table}[!t]
\caption{Comparative Evaluation of Centerline and Road Topology Performance for Offset Refinement Mechanisms.}
\label{table: proposal_configurations}
\centering
\scalebox{0.7}{
\begin{tabular}{lcccc}
\toprule
\textbf{Configuration} & \textbf{DET\textsubscript{l}} & \textbf{DET\textsubscript{l\_ch}} & \textbf{TOP\textsubscript{ll}} & \textbf{OLS\textsubscript{l}} \\
\midrule
Baseline (No Prediction)       & 31.1 & 31.7 & 22.5 & 36.8 \\
Single Proposal     & \underline{31.2} & \underline{37.6} & \underline{22.9} & \underline{38.9} \\
Multiple Proposals  & \textbf{33.1} & \textbf{37.9} & \textbf{25.0} & \textbf{40.3} \\
\bottomrule
\end{tabular}
}
\end{table}

A comparative analysis of two refinement strategies (introduced in Section~\ref{sec: proposal_mechanisms} and Section~\ref{sec: height_prediction}) is presented in Table~\ref{table: proposal_configurations}. The \textit{Baseline (No Prediction)} represents the coarse, discretized output from the standard expectation step (Eqs.~\eqref{eq: row_expectation}–\eqref{eq: col_expectation}), without any offset or height correction. In the \textit{Single Proposal} setting, the predicted offset and height values are applied \textbf{only} to these coarse, initially-extracted centerpoints. In contrast, in the \textit{Multiple Proposals} configuration, the predicted offset and height are applied to \textbf{every} pixel within the full rasterized mask region.

Both proposal mechanisms significantly outperform the baseline, confirming the effectiveness of the offset and height prediction heads. The \textit{Multiple Proposals} approach achieves the highest scores across all metrics, indicating that dense, non-directional refinement across the full mask region more robustly corrects discretization errors.

\subsection{Ablation Study of the Multi-Proposal}

\begin{table}[!t]
\caption{Ablation study on the individual and combined contributions of the proposed offset and height prediction heads.}
\label{table: multiple_proposal_ablation_study}
\centering
\scalebox{0.7}{
\begin{tabular}{lcccc}
\toprule
\textbf{Configuration} & \textbf{DET\textsubscript{l}} & \textbf{DET\textsubscript{l\_ch}} & \textbf{TOP\textsubscript{ll}} & \textbf{OLS\textsubscript{l}} \\
\midrule
No Prediction           & 31.1 & 31.7 & 22.5 & 36.8 \\
Only Offset             & 32.5 & 33.1 & \underline{23.8} & 38.2 \\
Only Height             & \underline{32.6} & \underline{37.2} & 23.4 & \underline{39.4} \\
Offset + Height         & \textbf{33.1} & \textbf{37.9} & \textbf{25.0} & \textbf{40.3} \\
\bottomrule
\end{tabular}
}
\end{table}

To analyze the effects of offset and height predictions in the multiple-proposal setup, we ran an ablation in which each component was enabled alone and together. Table~\ref{table: multiple_proposal_ablation_study} shows that both offset and height predictions contribute positively to centerline detection and road topology metrics. Notably, height prediction alone yields a more substantial improvement than offset prediction, particularly in DET\textsubscript{l\_ch}, suggesting it has a stronger effect on spatial localization. The combination of both produces the best overall results, indicating that offset and height predictions provide complementary benefits.

\subsection{Architectural Enhancements to the Mask Head}

\begin{table*}[!t]
\caption{OLS\textsubscript{l} for Bezier, Mask, and Fusion across splits (Original, Near), sensors (Camera, Cam+LiDAR), and ranges ($\pm 50$ m, $\pm 100$ m).}
\label{table: sensor_modalities_long_range}
\centering
\scalebox{0.7}{
\begin{tabular}{lcccccccc}
\toprule
\multirow{3}{*}{\textbf{Output Type}} & \multicolumn{4}{c}{\textbf{Original Split}} & \multicolumn{4}{c}{\textbf{Near Split}} \\
\cmidrule(lr){2-5} \cmidrule(lr){6-9}
& \multicolumn{2}{c}{Camera} & \multicolumn{2}{c}{Cam+LiDAR} & \multicolumn{2}{c}{Camera} & \multicolumn{2}{c}{Cam+LiDAR} \\
\cmidrule(lr){2-3} \cmidrule(lr){4-5} \cmidrule(lr){6-7} \cmidrule(lr){8-9}
& \([-50,+50]\) & \([-100,+100]\) & \([-50,+50]\) & \([-100,+100]\) & \([-50,+50]\) & \([-100,+100]\) & \([-50,+50]\) & \([-100,+100]\) \\
\midrule
\textbf{Bezier} & \textbf{43.5} & \underline{32.5} & \textbf{51.6} & \textbf{50.0} & \underline{27.8} & 16.5 & \textbf{32.0} & \underline{24.2} \\
\textbf{Mask}   & 40.8 & 31.0 & 47.7 & 46.9 & 26.4 & \underline{16.7} & \underline{31.0} & 23.7 \\
\textbf{Fusion} & \underline{42.5} & \textbf{32.9} & \underline{50.0} & \underline{48.6} & \textbf{27.9} & \textbf{17.4} & \textbf{32.0} & \textbf{24.5} \\
\bottomrule
\end{tabular}
}
\\ \footnotesize {Formatting: \textbf{bold} = best, \underline{underline} = second-best.}
\vspace{-0.5cm}
\end{table*}

\begin{table}[!t]
\caption{Ablation of mask-head architectural enhancements. Results are reported on the original split.}
\label{table: mask_head_enhancements}
\centering
\scalebox{0.7}{
\begin{tabular}{lcccc}
\toprule
\textbf{Method} & \textbf{DET\textsubscript{l}} & \textbf{DET\textsubscript{l\_ch}} & \textbf{TOP\textsubscript{ll}} & \textbf{OLS\textsubscript{l}} \\
\midrule
Base Mask Head                  & 33.3 & 38.3 & 24.6 & 40.4 \\
+ Auxiliary Bezier Regression   & 32.9 & 37.6 & 24.9 & 40.2 \\
+ ML1M (from MM)                & \underline{33.9} & \underline{38.9} & \underline{25.6} & \underline{41.1} \\
+ BDA (from MA)                 & \textbf{34.2} & \textbf{39.4} & \textbf{25.9} & \textbf{41.5} \\
\bottomrule
\end{tabular}
}
\end{table}

We analyze architectural modifications to the mask head, including auxiliary supervision,
matching, and attention. Table~\ref{table: mask_head_enhancements} reports the base mask
head and enhanced variants. MM is the baseline matcher; ML1M augments the bipartite
matching cost with a Bezier L1 term~\cite{li_mask_2023}; and BDA replaces MA foreground
attention~\cite{cheng_masked-attention_2022} with curve-aware attention around predicted
Bezier control points~\cite{kalfaoglu_topobda_2024}. Formal definitions are deferred to
Supplementary Sections~\ref{sup_sec: mask_l1_mix_matcher}
and~\ref{sup_sec: masked_and_bezier_deformable_attention_recap}.


The analysis reveals a key interaction: while the auxiliary Bezier regression head \textit{alone} slightly degrades performance (40.2 OLS\textsubscript{l} vs 40.4), it is a \textit{prerequisite} for the Mask-L1 Mix Matcher (ML1M). Enabling ML1M provides a net performance gain (41.1 OLS\textsubscript{l}), showing the matcher's benefit outweighs its dependency's minor interference. Since both ML1M and its auxiliary head are used only during training, this performance boost is achieved with no added inference cost.

Further improvements are achieved by integrating Bezier Deformable Attention (BDA) in place of standard mask attention. BDA enhances spatial modeling and yields the best overall performance. However, unlike ML1M, BDA requires Bezier predictions not only in training but also in inference. A detailed comparative ablation showing their impact across all three head types (Mask, Bezier, and Fusion) is provided in Section~\ref{sup_sec: comparative_ablation_of_architectural_enhancements_across_head_types}.


\subsection{Performance Across Output Types Under Sensor Modalities and Long-Range Conditions}
\label{sec: sensor_modalities_long_range}

We evaluated OLS\textsubscript{l} for Bezier, Mask, and Fusion outputs across sensor setups (camera-only vs. camera+LiDAR), spatial ranges ($\pm 50\text{ m}$ vs. $\pm 100\text{ m}$), and two dataset splits: \textbf{Original} (geographically overlapping) and \textbf{Near} (geographically disjoint). According to Table~\ref{table: sensor_modalities_long_range}, the key findings are:
\begin{itemize}[leftmargin=*,nosep]
\item
In the \textbf{Original} split, the Bezier head has the best OLS\textsubscript{l} in three of four configurations; the Fusion head only slightly outperforms it in the long-range camera-only case.

\item In the \textbf{Near} split, Fusion consistently scores highest across all configurations. The Mask head also outperforms Bezier in the long-range camera-only setting and shows consistently smaller Original$\rightarrow$Near degradation across all four matched settings, by 0.8 percentage points in the short-range camera-only case and by 2.1--3.1 points in the remaining cases, indicating stronger robustness to geographic shift.

\item Extending the range to $\pm 100\text{ m}$ causes large performance drops,
especially for camera-only models. In the \textbf{Near} split, the Fusion head drops
by 37.6\% (27.9$\rightarrow$17.4) for camera-only but by 23.4\%
(32.0$\rightarrow$24.5) with Cam+LiDAR, confirming that LiDAR provides the greatest
relative benefit at long range.

\item A larger relative LiDAR gain (Cam+LiDAR over camera-only OLS\textsubscript{l})
is observed in the \textbf{Original} split than in \textbf{Near}. For the Fusion head,
the gain is 17.6\% (Original) vs. 14.7\% (Near) at $\pm 50\text{ m}$, and
47.7\% (Original) vs. 40.8\% (Near) at $\pm 100\text{ m}$, indicating that
geographic overlap amplifies the apparent LiDAR benefit.

\end{itemize}

Together, these findings indicate that Bezier benefits more from overlap, whereas Mask and Fusion generalize more reliably across splits, sensors, and extended ranges. A detailed breakdown of these sensor fusion gains is provided in Section~\ref{sup_sec: sensor_fusion_impact_different_splits_ranges}.

\subsection{Performance Analysis Across Dataset Splits}
\label{sec: output_types_dataset_splits}

\begin{table}[!t]
\caption{OLS\textsubscript{l} generalization of Bezier, Mask, and Fusion across geographic splits in the standard camera-only, $\pm 50$ m setting: Original (overlap), Near, FarA, FarB, and FarC (disjoint).}
\label{table: split_wise_analysis}
\centering
\scalebox{0.7}{
\begin{tabular}{lccccc}
\toprule
\textbf{Output Type} & \textbf{Original} & \textbf{Near} & \textbf{FarA} & \textbf{FarB} & \textbf{FarC} \\
\midrule
Bezier               & \textbf{43.4} & 27.8 & 22.2 & \textbf{20.9} & 27.8 \\
Mask                 & 40.8 & 26.4 & 21.2 & 20.0 & 27.1 \\
Fusion               & \underline{42.5} & \textbf{27.9} & \textbf{22.3} & \underline{20.7} & \textbf{28.3} \\
\bottomrule
\end{tabular}
}
\\ \footnotesize {Formatting: \textbf{bold} = best, \underline{underline} = second-best.}
\end{table}

To study generalization across dataset distributions, we evaluated Bezier, Mask, and Fusion heads on five splits: \textbf{Original}, \textbf{Near}, \textbf{FarA}, \textbf{FarB}, and \textbf{FarC}, reporting OLS\textsubscript{l} scores at the standard range $\pm 50\text{ m}$ (Table~\ref{table: split_wise_analysis}). The key findings are:
\begin{itemize}[leftmargin=*,nosep]
\item Moving from the overlap-heavy \textbf{Original} split to geographically disjoint splits causes a severe performance drop for all output types. Averaging the per-split percentage drops over \textbf{Near}, \textbf{FarA}, \textbf{FarB}, and \textbf{FarC} gives relative OLS\textsubscript{l} reductions of 43.1\% for Bezier, 42.0\% for Mask, and 41.6\% for Fusion, quantifying the extent of benchmark inflation in the standard split.

\item While the Bezier head achieves the highest score on the \textbf{Original} split, the Fusion head outperforms it in three of the four geographically distinct splits: \textbf{Near}, \textbf{FarA}, and \textbf{FarC}; \textbf{FarB} remains nearly tied (20.9 vs. 20.7). The Mask head also shows consistently smaller per-split degradation than Bezier, by 0.7--2.4 percentage points across the four disjoint splits, further supporting stronger robustness under geographic shift.
\end{itemize}

This apparent Bezier advantage on the \textbf{Original} split also depends on coupling Bezier regression with BDA: as shown in \textbf{Supplementary Section~\ref{sup_sec: comparative_ablation_of_architectural_enhancements_across_head_types}}, the base Bezier head (with MA) is weaker than the Mask head, and fusion remains beneficial. This indicates that the Original-split advantage is not an inherent property of Bezier regression alone.

Taken together, severe performance drops observed for all output types on geographically disjoint splits show that the standard Original benchmark is inflated by geographic overlap, while more reliable generalization is exhibited by the Mask and Fusion heads.



\subsection{Comparison with SOTA in OpenLane-V2}
\label{sec: comparison_sota}

We adopt the geographically disjoint \textbf{Near} split as the primary benchmark for SOTA
comparison, with all competing methods re-trained on this split. As established in
Sections~\ref{sec: sensor_modalities_long_range} and \ref{sec: output_types_dataset_splits},
the standard \textbf{Original} split is afflicted by geographic data leakage that
disproportionately benefits Bezier-based representations. Crucially, this
evaluation employs the \textbf{score remapping} technique\footnote{Using score remapping
($P(x) \rightarrow P(x) + 1 \times [P(x) > 0.05]$). See Sec.~\ref{sec: experimental_evaluation}
and Sup.~Sec.~\ref{sup_sec: topology_metric} for details.} \cite{kalfaoglu_topomaskv2_2024}
to correct the 0.5 thresholding flaw in the standard TOP\textsubscript{ll} metric.

\begin{table}[!t]
\caption{Comparative evaluation on the \textbf{Near} split (\textbf{geographically disjoint}) using the V1.1 metric with \textbf{score remapping} applied to the TOP\textsubscript{ll} scores of all methods.}
\label{table: near_split_remapped}
\centering
\scalebox{0.70}{
\begin{tabular}{lcccc}
\toprule
\textbf{Method} & \textbf{DET\textsubscript{l}} & \textbf{DET\textsubscript{l\_ch}} & \textbf{TOP\textsubscript{ll}} & \textbf{OLS\textsubscript{l}} \\
\midrule
TopoNet \cite{li_graph-based_2023}      & 18.9 & 23.5 & 12.7 & 26.0 \\
TopoMLP \cite{wu_topomlp_2024}          & 15.6 & 22.4 & 14.5 & 25.3 \\
TopoLogic \cite{fu_topologic_2024}    & 16.9 & 22.7 & \textbf{15.5} & 26.3 \\
TopoMaskV2 (M) \cite{kalfaoglu_topomaskv2_2024} & 16.4 & 20.1 & 10.9 & 23.2 \\
TopoMaskV2 (F) \cite{kalfaoglu_topomaskv2_2024} & 18.5 & 23.8 & 11.7 & 25.5 \\
TopoBDA \cite{kalfaoglu_topobda_2024} & \textbf{20.8} & 24.9 & 13.0 & \underline{27.3} \\
\textbf{TopoMaskV3 (M) (Ours)}  & 19.3 & \underline{25.6} & 13.6 & \underline{27.3} \\
\textbf{TopoMaskV3 (F) (Ours)}  & \underline{20.5} & \textbf{26.2} & \underline{15.1} & \textbf{28.5} \\
\bottomrule
\end{tabular}
}
\\ \footnotesize {(M): Mask-based, and (F): Fusion-based approaches.}
\\ \footnotesize {Formatting: \textbf{bold} = best, \underline{underline} = second-best.}
\end{table}

As shown in Table~\ref{table: near_split_remapped}, \textbf{TopoMaskV3~(F)} achieves the
\textbf{state-of-the-art} OLS\textsubscript{l} score of \textbf{28.5} on the geographically
disjoint Near split, surpassing all competing methods including the strong TopoBDA baseline
(27.3 OLS\textsubscript{l}). The standalone \textbf{TopoMaskV3~(M)} also reaches
27.3 OLS\textsubscript{l}, matching TopoBDA and further confirming the generalization
strength of the mask-centric paradigm. A detailed breakdown of the score differences between
the flawed V1.1 metric and the stable, remapped metric is provided in Supplementary
Section~\ref{sup_sec: score_remapping_analysis}
(see Table~\ref{table: near_split_remapped_comparison} for a direct comparison).

For completeness and comparability with prior literature, a full SOTA comparison on the
standard \textbf{Original} split is provided in Supplementary
Section~\ref{sup_sec: sota_original_split}
(Table~\ref{tab: sota_camera_only_subsetA})\footnote{The standard OLS metric is reported
there, as OLS\textsubscript{l} cannot be computed for all methods since
DET\textsubscript{l\_ch} is not provided in most prior works.}.
On the Original split, \textbf{TopoMaskV3~(F)} ranks second with 50.1 OLS, behind
TopoBDA~\cite{kalfaoglu_topobda_2024} (51.7 OLS) --- a result directly consistent with the
Bezier memorization effect: TopoBDA, as a purely Bezier-based architecture, benefits
disproportionately from geographic overlap in the training data, artificially inflating its
apparent advantage --- an advantage that vanishes entirely on the geographically disjoint Near
split, where \textbf{TopoMaskV3~(F)} takes the lead.

%% file: topomaskv3/conclusion.tex
\section{Conclusion}
\label{sec: conclusion}
\vspace{-0.1cm}
This work introduced TopoMaskV3, which matures the mask-based paradigm for road topology
by incorporating dense offset and height prediction heads. On the geographically disjoint
Near split, \textbf{TopoMaskV3~(F)} achieves state-of-the-art \textbf{28.5~OLS\textsubscript{l}},
and the mask-only variant matches TopoBDA at 27.3 OLS\textsubscript{l}. We also introduced
geographically distinct splits and a long-range benchmark, showing that the mask head is
more robust to geographic overfitting than the Bezier head, while LiDAR fusion is most
beneficial at extended range and appears partially inflated on overlapping splits due to
memorization. Although TopoMaskV3 is not yet fully end-to-end, its dense offset field offers
a principled raster-specific alternative complementary to parametric approaches.

{\footnotesize
\noindent\textbf{Acknowledgements.}
We acknowledge computational resources from TRUBA (TÜBİTAK ULAKBİM) and the EuroHPC
Joint Undertaking (via MareNostrum 5 at BSC-CNS).
}

%% file: topomaskv3/supplementary.tex
\twocolumn[{%
    \centering
    \Large \textbf{Supplementary Material for: \\ TopoMaskV3: 3D Mask Head with Dense Offset and Height Predictions for Road Topology Understanding} \\[1.5em]
    \vspace{1em}
}]

\renewcommand{\thesection}{S}
\section{Supplementary Materials}
\beginsupplement

\subsection{Masked and Bezier Deformable Attention (Recap)}
\label{sup_sec: masked_and_bezier_deformable_attention_recap}
\vspace{-0.2cm}

\begin{figure}[tb]
  \centering
  \includegraphics[width=0.85\linewidth]{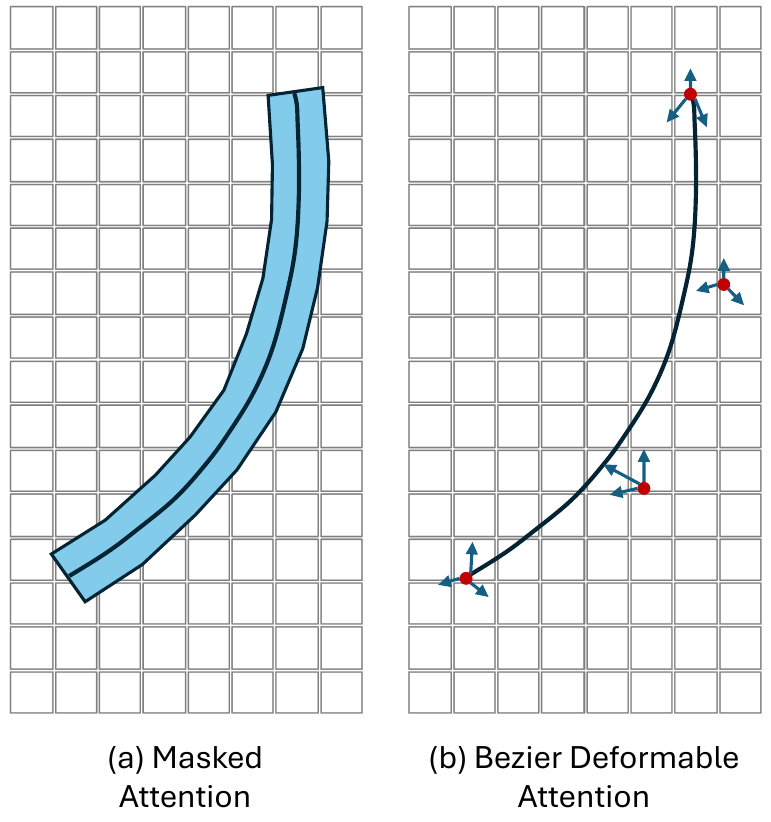}
  \caption{\textbf{Attention Mechanisms.} (a) Masked Attention restricts attention to foreground regions using a binary mask. (b) Bezier Deformable Attention (BDA) attends around predicted Bezier control points, enabling flexible and structure-aware feature aggregation.}
  \label{fig:masked_bezier_attention}
\end{figure}

\textbf{Masked Attention.} To focus attention on relevant regions, Masked Attention introduces a mask term $\mathbfcal{M}_{l-1}$ into the standard attention formulation:
\[
\mathbf{X}_{l} = \text{softmax}(\mathbfcal{M}_{l-1} + \mathbf{Q}_{l}\mathbf{K}^{\text{T}}_{l})\mathbf{V}_{l} + \mathbf{X}_{l-1}
\]
Here, $\mathbfcal{M}_{l-1}(x,y)=0$ for foreground pixels and $-\infty$ otherwise, effectively suppressing background attention. Masks are obtained by thresholding predicted probabilities at 0.5. For detailed information, please refer to the original Mask2Former study \cite{cheng_masked-attention_2022}. 

\textbf{Bezier Deformable Attention.} BDA extends deformable attention by replacing single reference points with Bezier control points $\mathbf{C}=\{\mathbf{c}_0,\dots,\mathbf{c}_N\}$:
\[
\text{BDA}(\mathbf{q},\mathbf{V},\mathbf{C})=\sum_{n=0}^{N}\sum_{k=1}^{K}A_{n,k}\mathbf{W}_n\mathbf{V}(\mathbf{c}_n+\Delta\mathbf{p}_{n,k})
\]
Each control point acts as an attention head, allowing the model to capture curve geometry without converting control points into dense polylines, reducing overhead and improving structural awareness. For detailed formulations and ablation studies, please refer to the original TopoBDA study \cite{kalfaoglu_topobda_2024}.

\subsection{Mask-L1 Mix Matcher (Recap)}
\label{sup_sec: mask_l1_mix_matcher}

The bipartite matching cost combines regression, mask, and classification terms:
\[
\mathcal{L}_{\text{l}} = \lambda_{\text{reg}} \mathcal{L}_{\text{reg}} + \lambda_{\text{mask}} \mathcal{L}_{\text{mask}} + \lambda_{\text{cls}} \mathcal{L}_{\text{cls}}.
\]
When $\lambda_{\text{reg}}=0$, the matcher relies on mask similarity; when $\lambda_{\text{mask}}=0$, it becomes an L1-based matcher. Activating both terms yields the \textit{Mask-L1 Mix Matcher}, which follows the principle of MaskDINO \cite{li_mask_2023} by jointly leveraging regression and mask losses in the matching cost.


\subsection{Detailed Curve Reconstruction Pipeline}
\label{sup_sec: curve_reconstruction_details}
The set of refined 3D points $\{\tilde{\mathbf{p}}_{grid}\}$ generated by the offset proposal (Section~\ref{sec: proposal_mechanisms}) and height-estimation (Section~\ref{sec: height_prediction}) steps is initially in the BEV grid's pixel coordinate system. These points must be transformed and regularized into a smooth, ordered, continuous curve in real-world coordinates. This multi-step process is guided by the \textbf{quad-direction label} predicted by the classification head.

\subsubsection{Coordinate Transformation}
Each refined grid point $\tilde{\mathbf{p}}_{grid} = (i, j, h)$ (where $(i,j)$ is the refined 2D location and $h$ is the estimated normalized height value) is first mapped to its real-world 3D coordinate $\tilde{\mathbf{p}}_{world} = (x, y, z)$ via a pre-defined grid-to-world transformation matrix $\mathbf{V}^{-1}$. This step yields a set of unordered 3D points $\{\tilde{\mathbf{p}}_{world}\}$ in the real-world space, where $x$ represents the vertical (forward) axis and $y$ represents the horizontal (lateral) axis.

\subsubsection{Direction-Aware Regularization and Ordering}
This set of noisy, real-world points $\{\tilde{\mathbf{p}}_{world}\}$ is processed to form the final smooth curve. This regularization, which involves joint polynomial fitting and arc-length interpolation, is crucial for further smoothing discretization artifacts. The benefits of this approach are experimentally evaluated in Section~\ref{sup_sec: post_process_impact}.

The process involves three main operations:
\paragraph{Polynomial Functions Fit}
First, polynomial functions are fit to the unordered 3D points $\{\tilde{\mathbf{p}}_{world}\}$. This finds the best-fit curves that represent the underlying shape of the noisy points. Two separate least-squares fits are performed:
\begin{itemize}[nosep]
    \item \textbf{2D Path Fit:} A direction-aware polynomial is fit to the $(x, y)$ coordinates based on the quad-direction label such that $y=f(x)$ for `up'/`down' directions and $x=f(y)$ for `left'/`right' directions.
    
    \item \textbf{3D Height Fit:} A 2D polynomial surface, $z = g(x, y)$, is fit to the 3D points $(x, y, z)$ by solving for the coefficients $\mathbf{C}$ of a 3rd-order polynomial:
    \begin{equation}
        z \approx g(x,y) = C_0 + \sum_{i=1}^{3} (C_{2i-1} x^i + C_{2i} y^i).
    \end{equation}
\end{itemize}

\paragraph{3D Curve Generation and Resampling}
Second, the 3D curve is generated and resampled. A preliminary 3D polyline, $\mathcal{P}_{poly}$, is generated by sampling the fitted 2D path $f$ at linearly-spaced intervals (e.g., using `linspace`) to get new $(x_i, y_i)$ points. The 3D height surface $g$ is then evaluated at these locations to get the corresponding $z_i$ coordinates. This $\mathcal{P}_{poly}$ is smooth but its points are not equidistant. This 3D polyline is then resampled using \textbf{arc-length interpolation} (`interp\_arc`), which takes $\mathcal{P}_{poly}$ as input and produces the final, uniformly sampled set of $N$ equidistant 3D points, $\{\tilde{\mathbf{p}}'\} = \{(x'_i, y'_i, z'_i)\}_{i=1}^N$.

\paragraph{Final Sorting}
Finally, this resulting set $\{\tilde{\mathbf{p}}'\}$ is \textbf{explicitly sorted} to guarantee the correct directional flow, as defined by the quad-direction label. For instance, points for an 'up' centerline are sorted by increasing $x'$-coordinates, while 'down' centerlines are sorted by decreasing $x'$-coordinates. This step ensures the final point sequence is correctly ordered from start to end.

\subsection{LiDAR Fusion Pipeline}
\label{sup_sec: lidar_fusion_details}
The sensor fusion pipeline integrates camera and LiDAR data at the voxel level before projecting to a unified BEV representation. This early fusion in 3D space preserves fine-grained spatial information. The process, adapted from~\cite{kalfaoglu_topobda_2024}, is as follows:

\subsubsection{Camera Feature Voxelization}
A set of $N$ perspective-view images $\{\mathbf{I}_i\}_{i=1}^N$ is first processed by a backbone $f_{PV}$ to extract features $\{\mathbf{F}_{PV_i}\}_{i=1}^N$. These 2D features are then lifted into a 3D voxel representation $\mathbf{F}_{\text{voxelCam}} \in \mathbb{R}^{H_{\text{bev}} \times W_{\text{bev}} \times Z \times C_{\text{camera}}}$ using a voxelization module $f_{voxel}$ (e.g., Lift-Splat~\cite{philion_lift_2020}):
\begin{equation}
    \mathbf{F}_{\text{voxelCam}} = f_{voxel}(\{\mathbf{F}_{PV_i}\}_{i=1}^N).
\end{equation}

\subsubsection{LiDAR Feature Voxelization}
Concurrently, the raw LiDAR point cloud $\{\mathbf{p}_{\text{lidar}}^j\}_{j=1}^{N_{\text{lidar}}}$ is processed by a dedicated LiDAR encoder $f_{\text{lidar}}$ (e.g., SECOND~\cite{yan_second_2018}). This module converts the sparse point cloud into a dense, feature-rich voxel grid $\mathbf{F}_{\text{voxelLidar}} \in \mathbb{R}^{H_{\text{bev}} \times W_{\text{bev}} \times Z \times C_{\text{lidar}}}$ that shares the same spatial dimensions:
\begin{equation}
    \mathbf{F}_{\text{voxelLidar}} = f_{\text{lidar}}(\{\mathbf{p}_{\text{lidar}}^j\}_{j=1}^{N_{\text{lidar}}}).
\end{equation}

\subsubsection{Voxel-Space Fusion}
The two feature-rich voxel grids are concatenated along the channel dimension to create a unified fused tensor, $\mathbf{F}_{\text{fused}} \in \mathbb{R}^{H_{\text{bev}} \times W_{\text{bev}} \times Z \times (C_{\text{camera}} + C_{\text{lidar}})}$:
\begin{equation}
    \mathbf{F}_{\text{fused}} = \text{concat}(\mathbf{F}_{\text{voxelCam}}, \mathbf{F}_{\text{voxelLidar}}).
\end{equation}

\subsubsection{BEV Feature Map Generation}
Finally, the height ($Z$) and channel dimensions of the fused voxels are flattened and passed through a 2D convolutional layer $f_{\text{conv2}}$ to produce the final, unified BEV feature map $\mathbf{F}_{\text{bev}} \in \mathbb{R}^{H_{\text{bev}} \times W_{\text{bev}} \times C_{\text{BEV}}}$:
\begin{equation}
    \mathbf{F}_{\text{bev}} = f_{\text{conv2}}(\mathbf{F}_{\text{fused}}).
\end{equation}
This BEV map serves as the input to the downstream decoder heads.

\subsection{Topology Metric}
\label{sup_sec: topology_metric}
The $\text{TOP}_{ll}$ metric is an Average Precision (AP)-based measure for evaluating directed lane-to-lane topology in graph-structured predictions. A predicted adjacency matrix $P_\theta$ is constructed and aligned with the full ground-truth (GT) vertex set $V$ by matching predicted and GT vertices using Fr\'{e}chet distance thresholds $\Theta = \{1\,\text{m}, 2\,\text{m}, 3\,\text{m}\}$.

For each GT edge:
\begin{itemize}[nosep]
    \item If both vertices are matched, the corresponding entry in $P_\theta$ is set to the model's predicted edge confidence.
    \item If either vertex is unmatched, the entry is set to $0$ for GT-positive edges and to $0.5 + \varepsilon$ for GT-negative edges. This is intended to penalize missing detections as low-confidence false positives.
\end{itemize}




For each GT vertex $v \in V$, AP is computed independently for outgoing ($d=\text{out}$) and incoming ($d=\text{in}$) edges. The AP for a given vertex, threshold $\theta$, and direction $d$ is:
\begin{equation}
    \text{AP}(v, \theta, d) = \frac{1}{|N_d(v)|} \sum_{\hat{n}' \in \hat{N}'_{\theta, d}(v)} P_{\theta, d}(\hat{n}') \mathbb{I}(\hat{n}' \in N_d(v))
    \label{eq:ap_sub}
\end{equation}
where $N_d(v)$ is the set of ground-truth neighbors, $\hat{N}'_{\theta, d}(v)$ is the ranked predicted neighbor list, $P_{\theta, d}(\hat{n}')$ is the precision at that rank, and $\mathbb{I}$ is the indicator function.

The final $\text{TOP\textsubscript{ll}}$ score is the mean of this AP over all thresholds, all vertices, and both directions:
\begin{equation}
    \text{TOP\textsubscript{ll}} = \frac{1}{2|\Theta||V|} \sum_{\theta \in \Theta} \sum_{v \in V} \sum_{d \in \{\text{in, out}\}} \text{AP}(v, \theta, d)
    \label{eq:top_ll}
\end{equation}

\vspace{0.3em}
\noindent\textbf{Critical Analysis of the 0.5 Confidence Threshold.}
A significant flaw in the V1.1 metric implementation is that only predicted edges with \textbf{confidence greater than 0.5} are ranked for the AP calculation. This is a non-standard practice for an AP-based metric, which should evaluate the entire precision-recall curve by ranking all predictions. This 0.5 thresholding effectively ignores all predictions in the lower half of the confidence range.

As demonstrated in prior work~\cite{kalfaoglu_topomaskv2_2024}, this flaw can be exploited to artificially inflate scores. Simply remapping low-confidence predictions (e.g., $> 0.05$) to a value just above the 0.5 threshold can substantially boost performance (e.g., by +10.6 TOP\textsubscript{ll} for some models~\cite{kalfaoglu_topomaskv2_2024}) without any change to the model itself. This proves the metric, in its current form, is not a robust measure of performance. The quantitative impact of this remapping is detailed in our Supplementary Section~\ref{sup_sec: score_remapping_analysis}, which provides a direct comparison (Table~\ref{table: near_split_remapped_comparison}) between the ``flawed" V1.1 metric and the ``healthy" remapped scores used in our main paper's Table~\ref{table: near_split_remapped}.

A proper AP evaluation should rank all predictions. A simple and effective fix would be to lower this ranking threshold from $0.5$ to a near-zero value (e.g., \textbf{0.01}). Correspondingly, the penalty for unmatched GT-negative edges should be set to \textbf{$0.01 + \varepsilon$} to ensure they are correctly penalized as low-confidence false positives.

\subsection{Loss Functions}
\label{sup_sec: experiments_loss_functions}

The overall training objective is a multi-component loss function. The total loss $\mathcal{L}_{\text{total}}$ is a weighted sum of the primary centerline loss $\mathcal{L}_{\text{l}}$, a traffic element loss $\mathcal{L}_{\text{t}}$ (from DAB-DETR~\cite{liu_dab-detr_2022}), and topology losses $\mathcal{L}_{\text{ll}}$ and $\mathcal{L}_{\text{lt}}$ (from TopoNet~\cite{li_graph-based_2023}).

\subsubsection{Total Loss}
The total loss is defined as:
\begin{equation*}
    \mathcal{L}_{\text{total}} = \mathcal{L}_{\text{l}} + \mathcal{L}_{\text{t}} + \mathcal{L}_{\text{ll}} + \mathcal{L}_{\text{lt}}.
\end{equation*}
Our primary contributions are captured within the centerline loss $\mathcal{L}_{\text{l}}$, which is a composite of five distinct terms:
\begin{multline*}
    \mathcal{L}_{\text{l}} = \lambda_{\text{cls}} \mathcal{L}_{\text{cls}} + \lambda_{\text{reg}} \mathcal{L}_{\text{reg}} + \lambda_{\text{mask}} \mathcal{L}_{\text{mask}} \\
    + \lambda_{\text{offset}} \mathcal{L}_{\text{offset}} + \lambda_{\text{height}} \mathcal{L}_{\text{height}}.
\end{multline*}

\subsubsection{Core Centerline Losses}
Three components of the centerline loss are adapted from prior work.

\paragraph{Classification Loss ($\mathcal{L}_{\text{cls}}$)}
A standard cross-entropy loss is applied to each query to predict its class. This includes the four quad-direction labels (as explained in Section \ref{sec: methodology}) and a "no-object" class for empty queries. Due to the dominance of empty queries, the "no-object" class is down-weighted by a factor of 0.1 following \cite{cheng_masked-attention_2022}.

\paragraph{Bezier Regression Loss ($\mathcal{L}_{\text{reg}}$)}
For the optional Bezier head, a standard L1 regression loss is applied between the $N$ predicted normalized control points $\mathbf{c}_i$ and the ground-truth points $\hat{\mathbf{c}}_i$:
\begin{equation*}
    \mathcal{L}_{\text{reg}} = \frac{1}{L} \sum_{j=1}^{L} \sum_{i=0}^{N} \| \mathbf{c}_{i,j} - \hat{\mathbf{c}}_{i,j} \|_1,
\end{equation*}
where $L$ is the number of matched ground-truth centerlines.

\paragraph{Instance Mask Loss ($\mathcal{L}_{\text{mask}}$)}
Following~\cite{cheng_masked-attention_2022}, the instance mask loss is a combination of a Binary Cross Entropy (BCE) loss and a Dice loss. This is computed efficiently by sampling $K$ points $\{\mathbf{a}_k\}$ from the predicted probability map $\mathbf{M}_{\text{prob}}$. The total mask loss is the sum of $\mathcal{L}_{\text{BCE}}$ and $\mathcal{L}_{\text{Dice}}$:
\begin{align*}
    \mathcal{L}_{\text{BCE}} &= \frac{1}{K} \sum_{k=1}^K \text{BCE}(\mathbf{M}_{\text{prob}}(\mathbf{a}_k), \mathbf{G}_{\text{map}}(\mathbf{a}_k)) \\
    \mathcal{L}_{\text{Dice}} &= 1 - \frac{2 \sum_{k=1}^K \mathbf{M}_{\text{prob}}(\mathbf{a}_k) \mathbf{G}_{\text{map}}(\mathbf{a}_k)}{\sum_{k=1}^K \mathbf{M}_{\text{prob}}(\mathbf{a}_k) + \sum_{k=1}^K \mathbf{G}_{\text{map}}(\mathbf{a}_k)}
\end{align*}

\subsubsection{Dense Offset and Height Losses (Ours)}
To supervise the new architectural components introduced in the main paper (Sections~\ref{sec: proposal_mechanisms} and \ref{sec: height_prediction}), two novel L1-based losses are introduced.

A key aspect of this supervision is the definition of the foreground mask, $\mathbf{M}_{\text{fg}}$. For each pixel $(i,j)$, the L2 norm (Euclidean distance) of its target offset vector $\hat{\mathbf{O}}(i,j)$ is computed. The pixel is included in the foreground mask only if this norm is less than a centerline thickness threshold, which is set to 4 pixels.
\begin{equation*}
    \mathbf{M}_{\text{fg}}(i,j) = \mathbb{I}\big[\|\hat{\mathbf{O}}(i,j)\|_2 < 4\big]
\end{equation*}
This defines a supervision band, capturing all grid pixels whose closest point on the continuous centerline is up to 4 pixels away.

\paragraph{Offset Loss ($\mathcal{L}_{\text{offset}}$)}
The dense offset head predicts an offset map $\mathbf{O} \in \mathbb{R}^{H \times W \times 2}$. This is supervised by a masked L1 loss against the target offset map $\hat{\mathbf{O}}$, using the dynamic foreground mask $\mathbf{M}_{\text{fg}}$. The loss is normalized by the number of active foreground pixels:
\begin{equation*}
    \mathcal{L}_{\text{offset}} = \frac{1}{\sum \mathbf{M}_{\text{fg}}} \sum_{i,j} \| \mathbf{O}(i,j) - \hat{\mathbf{O}}(i,j) \|_1 \cdot \mathbf{M}_{\text{fg}}(i,j).
\end{equation*}

\paragraph{Height Loss ($\mathcal{L}_{\text{height}}$)}
Similarly, the dense height head predicts a height map $\mathbf{H} \in \mathbb{R}^{H \times W}$. This is supervised by a masked L1 loss against the target height map $\hat{\mathbf{H}}$, using the exact same foreground mask $\mathbf{M}_{\text{fg}}$:
\begin{equation*}
    \mathcal{L}_{\text{height}} = \frac{1}{\sum \mathbf{M}_{\text{fg}}} \sum_{i,j} | \mathbf{H}(i,j) - \hat{\mathbf{H}}(i,j) | \cdot \mathbf{M}_{\text{fg}}(i,j).
\end{equation*}

\subsection{Implementation Details}
\label{sup_sec: implementation_details}

\subsubsection{Architecture Overview}
The model architecture is built upon the TopoMaskV2~\cite{kalfaoglu_topomaskv2_2024} and TopoBDA~\cite{kalfaoglu_topobda_2024} frameworks, utilizing distinct, non-weight-sharing backbones for the traffic element and centerline branches. The traffic element branch is based on DAB-DETR~\cite{liu_dab-detr_2022} and integrates concepts from DN-DETR~\cite{li_dn-detr_2022} and DINO~\cite{zhang_dino_2022}. For all experiments in this study, both branches use the ResNet50~\cite{maintainers_torchvision_2016} architecture.

For BEV feature generation, the multi-height bin Lift-Splat~\cite{philion_lift_2020, kalfaoglu_topomaskv2_2024} and efficient Voxel Pooling~\cite{huang_bevpoolv2_2022} implementations were used. The topology heads project the query embeddings from both branches into a shared space using MLPs, concatenate them, and process the result with a final MLP.

\subsubsection{Optimization Parameters}
The model was trained with a batch size of 8 using the AdamW optimizer. The base learning rate was set to \(3 \times 10^{-4}\) with a weight decay of \(1 \times 10^{-2}\). The learning rates for both the PV and BEV backbones were scaled by 0.1. A polynomial learning rate decay schedule (factor 0.9) with 1000 warm-up iterations was used. Gradient norm clipping was set to 35. As detailed in Section~\ref{sup_sec: experiments_loss_functions}, the loss coefficients for the bipartite matcher were: \(\lambda_{\text{cls}}=2\), \(\lambda_{\text{reg}}=5\), \(\lambda_{\text{mask}_{BCE}}=5\), \(\lambda_{\text{mask}_{Dice}}=5\), \(\lambda_{\text{offset}}=20\), and \(\lambda_{\text{height}}=50\).

\subsubsection{Architecture Hyperparameters}
The BEV grid dimensions ($H_{\text{bev}}$, $W_{\text{bev}}$) were set to 200 and 104, respectively, at a 0.5m x 0.5m resolution. The vertical dimension ($Z$) was set to 20 bins, spanning \([-10, 10]\) meters. For ground-truth generation, the centerline mask width was set to 4 pixels, following~\cite{kalfaoglu_topomaskv2_2024, kalfaoglu_topobda_2024}. When the Bezier branch was active, the number of control points was set to 4~\cite{wu_topomlp_2024, kalfaoglu_topobda_2024}. The number of queries was 200, and the transformer hidden channel dimension was 256. The transformer encoder and decoder were set to 6 and 10 layers, respectively. Multi-scale features were used from PV scales $\frac{1}{8}$, $\frac{1}{16}$, and $\frac{1}{32}$, and BEV scales $1$, $\frac{1}{2}$, and $\frac{1}{4}$.

\subsubsection{Dataset Preprocessing}
Standard preprocessing, including a uniform 0.5× scaling, was applied to all camera inputs. For Subset-A, images were resized to $1024 \times 736$ (width $\times$ height) with top crops of $178$ pixels (front) or $19$ pixels (others) after scaling by $0.5 \times$ .

\subsubsection{Benchmark Release}
\label{sup_sec: benchmark_release}
To support reproducibility of the Argoverse2-derived benchmark, we release the processed annotations, split definitions, prebuilt pkl files, and setup instructions through our OpenLane-V2 benchmark fork\footnote{\href{https://github.com/artest08/OpenLane-V2/tree/different_splits}{\nolinkurl{https://github.com/artest08/OpenLane-V2/tree/different_splits}}}. This release covers the \textbf{Original}, \textbf{Near}, \textbf{FarA}, \textbf{FarB}, and \textbf{FarC} protocols and their corresponding $\pm 100$ m variants. The current public release is centered on these benchmark artifacts and usage instructions rather than the full post-processing code.

\subsection{Comparative Ablation of Architectural Enhancements Across Head Types}
\label{sup_sec: comparative_ablation_of_architectural_enhancements_across_head_types}

\begin{table}[ht]
\caption{Comparative ablation of architectural enhancements across different head types, with results reported using OLS$_l$.}
\label{sup_table: architectural_enhancements_on_different_output_types}
\centering
\scalebox{0.7}{
\begin{tabular}{lccc}
\toprule
\textbf{Enhancement} & \textbf{Bezier} & \textbf{Mask} & \textbf{Fusion} \\
\midrule
Base                         & 38.3 & 40.2 & 40.6 \\
+ ML1M (from MM)             & \underline{39.0} & \underline{41.1} & \underline{41.5} \\
+ BDA (from MA)              & \textbf{43.8} & \textbf{41.5} & \textbf{42.8} \\
\bottomrule
\end{tabular}
}
\end{table}

To assess the impact of enhancements introduced in the TopoBDA study~\cite{kalfaoglu_topobda_2024} when adapted to our framework, we perform a comparative ablation across three head types: Bezier, Mask, and Fusion. Table~\ref{sup_table: architectural_enhancements_on_different_output_types} reports OLS\textsubscript{l} scores for the base configuration and two adapted mechanisms—Mask L1 Mix Matcher (ML1M) and Bezier Deformable Attention (BDA).

Although ML1M and BDA are not proposed in this study, they are adapted from the TopoBDA framework to evaluate their compatibility and effectiveness within our mask head design. ML1M yields consistent improvements across all head types, suggesting its general utility. In contrast, BDA shows a particularly strong effect on the Bezier head, likely due to the direct alignment between Bezier regression outputs and the Bezier deformable attention mechanism. This synergy may facilitate more effective gradient flow and feature refinement. While BDA also benefits the Mask and Fusion heads, its most pronounced impact is observed in the Bezier head. 


This strong synergistic effect of BDA on the Bezier head is pivotal. As shown in Table~\ref{sup_table: architectural_enhancements_on_different_output_types}, in the \textbf{Base} (MA) configuration, the Bezier head (38.3 OLS\textsubscript{l}) is weaker than the Mask head (40.2 OLS\textsubscript{l}), and Fusion (40.6 OLS\textsubscript{l}) provides a clear benefit. With the integration of \textbf{BDA}, this dynamic reverses: the Bezier head (43.8 OLS\textsubscript{l}) becomes the superior component, and Fusion (42.8 OLS\textsubscript{l}) is no longer beneficial. This explains why the enhanced standalone Bezier head outperforms the Mask and Fusion heads on the \textbf{Original} dataset split. However, this outcome is shown to vary on different data partitions, as a more detailed analysis across dataset splits is presented in the main paper Section~\ref{sec: output_types_dataset_splits}.

\subsection{Sensor Fusion Impact on Different Splits and Ranges}
\label{sup_sec: sensor_fusion_impact_different_splits_ranges}

\begin{table*}[t]
\caption{Detection performance across geographic splits, perception ranges, and sensor configurations. Metrics include DET\textsubscript{l}, DET\textsubscript{l\_ch}, TOP\textsubscript{ll}, and OLS\textsubscript{l}, with relative improvements from LiDAR integration reported as OLS\%. This analysis is conducted using the TopoBDA benchmark setup.}
\label{table: sensor_fusion_splits_ranges}
\centering
\scalebox{0.70}{
\begin{tabular}{l l l c c c c c}
\toprule
\textbf{Split} & \textbf{Range} & \textbf{Sensor} & \textbf{DET\textsubscript{l}} & \textbf{DET\textsubscript{l\_ch}} & \textbf{TOP\textsubscript{ll}} & \textbf{OLS\textsubscript{l}} & \textbf{OLS\%↑} \\
\midrule
Original & $\pm 50\text{ m}$ & RGB             & 37.6 & 37.7 & 28.3 & 42.9 & -- \\
Original & $\pm 50\text{ m}$ & RGB + LiDAR     & \textbf{47.0} & \textbf{49.8} & \textbf{37.0} & \textbf{52.6} & \textbf{22.61\%} \\
\midrule
Original & $\pm 100\text{ m}$ & RGB           & 24.1 & 28.5 & 20.4 & 32.6 & -- \\
Original & $\pm 100\text{ m}$ & RGB + LiDAR   & \textbf{43.4} & \textbf{48.4} & \textbf{34.7} & \textbf{50.2} & \textbf{53.99\%} \\
\midrule
Near     & $\pm 50\text{ m}$ & RGB             & 19.2 & 24.5 & 15.2 & 27.6 & -- \\
Near     & $\pm 50\text{ m}$ & RGB + LiDAR     & \textbf{24.2} & \textbf{31.6} & \textbf{18.0} & \textbf{32.7} & \textbf{18.48\%} \\
\midrule
Near     & $\pm 100\text{ m}$ & RGB           & 12.1 & 15.6 & 6.7  & 17.9 & -- \\
Near     & $\pm 100\text{ m}$ & RGB + LiDAR   & \textbf{18.4} & \textbf{22.5} & \textbf{12.2} & \textbf{25.3} & \textbf{41.34\%} \\
\bottomrule
\end{tabular}
}
\end{table*}

Table~\ref{table: sensor_fusion_splits_ranges} presents detection performance across different geographic splits, sensor configurations, and perception ranges. In this analysis, the TopoBDA \cite{kalfaoglu_topobda_2024} architecture has been utilized. Across all settings, incorporating LiDAR consistently improves detection metrics compared to using RGB alone. Notably, the performance gain from LiDAR becomes more pronounced in the extended range setting ($\pm 100\text{ m}$), where RGB-only setups suffer from significant degradation. 

In both range settings, LiDAR integration yields consistently higher relative improvements in the \textbf{Original} split compared to the \textbf{Near} split. Specifically, in the $\pm 50\text{ m}$ range, LiDAR provides a relative gain of \textbf{22.61\%} in the \textbf{Original} split, while the improvement in the \textbf{Near} split is only \textbf{18.48\%}. This trend becomes even more pronounced in the extended $\pm 100\text{ m}$ range, where the relative gain reaches \textbf{53.99\%} in the \textbf{Original} split versus \textbf{41.34\%} in the \textbf{Near} split. These discrepancies suggest that LiDAR-based models may benefit more from geographic overlap, potentially leveraging memorized spatial priors. In contrast, RGB-only models, while less performant overall, exhibit more stable generalization across unseen regions. These findings highlight the importance of evaluating sensor fusion strategies not only by absolute performance but also by their robustness across diverse geographic domains.

\subsection{Analysis of Score Remapping on the Near Split}
\label{sup_sec: score_remapping_analysis}

\begin{table}[ht]
\caption{Direct comparison of topology and OLS scores with and without score remapping on the Near Split. Note the significant, non-uniform boost from remapping.}
\label{table: near_split_remapped_comparison}
\centering
\scalebox{0.7}{
\begin{tabular}{lcccc}
\toprule
\multirow{2}{*}{\textbf{Method}} & \multicolumn{2}{c}{\textbf{Flawed Metric}} & \multicolumn{2}{c}{\textbf{Remapped Metric}} \\
\cmidrule(lr){2-3} \cmidrule(lr){4-5}
 & TOP\textsubscript{ll} & OLS\textsubscript{l} & TOP\textsubscript{ll} & OLS\textsubscript{l} \\
\midrule
TopoNet         & 6.2  & 22.4 & 12.7 \textcolor{blue}{(+6.5)} & 26.0 \textcolor{blue}{(+3.6)} \\
TopoMLP         & 13.0 & 24.7 & 14.5 \textcolor{blue}{(+1.5)} & 25.3 \textcolor{blue}{(+0.6)} \\
TopoLogic       & 15.0 & 26.1 & 15.5 \textcolor{blue}{(+0.5)} & 26.3 \textcolor{blue}{(+0.2)} \\
TopoMaskV2 (M)  & 6.1  & 20.4 & 10.9 \textcolor{blue}{(+4.8)} & 23.2 \textcolor{blue}{(+2.8)} \\
TopoMaskV2 (F)  & 6.6  & 22.6 & 11.7 \textcolor{blue}{(+5.1)} & 25.5 \textcolor{blue}{(+2.9)} \\
TopoBDA       & 10.6 & 26.1 & 13.0 \textcolor{blue}{(+2.4)} & 27.3 \textcolor{blue}{(+1.2)} \\
\textbf{TopoMaskV3 (M)} & 5.8 & 23.0 & 13.6 \textcolor{blue}{(+7.8)} & 27.3 \textcolor{blue}{(+4.3)} \\
\textbf{TopoMaskV3 (F)} & 6.6 & 24.2 & 15.1 \textcolor{blue}{(+8.5)} & 28.5 \textcolor{blue}{(+4.3)} \\
\bottomrule
\end{tabular}
}
\end{table}

As detailed in the main paper and Supplementary Section~\ref{sup_sec: topology_metric}, the standard TOP\textsubscript{ll} metric's 0.5 thresholding flaw prevents a stable comparison of topological reasoning. The main paper's \textbf{Section~\ref{sec: comparison_sota}} (Table~\ref{table: near_split_remapped}) presents the SOTA comparison on the `Near Split' using the corrected, ``healthy" score remapping.

To illustrate the impact of this correction, \textbf{Table~\ref{table: near_split_remapped_comparison}} in this section provides a direct side-by-side comparison of the OLS\textsubscript{l} and TOP\textsubscript{ll} scores computed with the \textbf{flawed} V1.1 metric versus the \textbf{remapped} metric.

Table~\ref{table: near_split_remapped_comparison} quantifies the impact of applying the score remapping technique ($P(x) \rightarrow P(x) + 1 \times [P(x) > 0.05]$) \cite{kalfaoglu_topomaskv2_2024}. As shown, this yields significantly higher and more representative TOP\textsubscript{ll} scores for most methods (e.g., TopoNet's score increases from 6.2 to 12.7). However, the impact is non-uniform across different methods. For \textbf{TopoLogic}, applying the remapping technique was not feasible, as it already implements its own inherent distance-based score manipulation. To avoid the complexity of compounding these two strategies, a comparable ``healthy" result was obtained by simply modifying the standard metric's evaluation threshold from 0.5 to 0.05 for its outputs. Other methods, such as TopoMLP, are also less affected, as their original implementations already produce high-confidence predictions that naturally bypass the flawed 0.5 threshold. This non-uniform impact highlights the instability of the original metric and validates the use of score remapping in the main paper (Table~\ref{table: near_split_remapped}) for a more stable and fair comparison.

\subsection{Impact of Post-Processing Methods on Instance-Level Point Outputs}
\label{sup_sec: post_process_impact}

\begin{table}[ht]
\caption{Performance comparison of post-processing methods. Best scores are bolded, second-best scores are underlined.}
\label{table:postprocess_analysis}
\centering
\scalebox{0.85}{
\begin{tabular}{lcccc}
\toprule
\textbf{Method} & \textbf{DET\textsubscript{l}} & \textbf{DET\textsubscript{l\_ch}} & \textbf{TOP\textsubscript{ll}} & \textbf{OLS\textsubscript{l}} \\
\midrule
None            & 28.1 & 34.0 & 17.6 & 34.7 \\
P2              & 30.6 & 37.1 & 21.9 & 38.2 \\
P3              & 32.7 & 37.4 & 23.4 & 39.5 \\
Arc             & 25.7 & 33.0 & 14.5 & 32.2 \\
P3+Arc          & \underline{33.1} & \underline{37.8} & \underline{25.0} & \underline{40.3} \\
P4+Arc          & \textbf{33.3} & \textbf{38.0} & \textbf{25.8} & \textbf{40.7} \\
\bottomrule
\end{tabular}
}
\end{table}

To evaluate the effect of different post-processing strategies applied to the proposed instance-level point outputs, several configurations were tested. These include polynomial fitting of varying degrees and arc interpolation. Polynomial fitting was applied in the vertical-horizontal domain using second-, third-, and fourth-order polynomials. Arc interpolation was also considered both independently and in combination with polynomial fitting. Table~\ref{table:postprocess_analysis} summarizes the results across four metrics: DET\textsubscript{l}, DET\textsubscript{l\_chamfer}, TOP\textsubscript{ll}, and OLS\textsubscript{l}.

As shown in Table~\ref{table:postprocess_analysis}, applying polynomial fitting significantly improves performance over the raw point outputs. While arc interpolation alone degrades performance, its combination with polynomial fitting—particularly third- and fourth-order—leads to notable gains. The best overall results are achieved with fourth-order polynomial fitting combined with arc interpolation, indicating that higher-order curve modeling and smooth interpolation are complementary in refining centerline predictions. This also highlights a limitation of the current mask-based formulation: unlike the direct Bezier path, it requires additional thresholding, dense refinement, and curve reconstruction after decoder prediction, and therefore introduces extra post-processing overhead.

\subsection{Threshold Sensitivity for Mask-Based Point Selection}
\label{sup_sec: threshold_value_impact_on_results}

\begin{table}[ht]
\caption{Performance across different mask probability thresholds. Best scores are bolded, second-best scores are underlined.}
\label{table:threshold_analysis}
\centering
\scalebox{0.85}{
\begin{tabular}{lcccc}
\toprule
\textbf{Threshold} & \textbf{DET\textsubscript{l}} & \textbf{DET\textsubscript{l\_ch}} & \textbf{TOP\textsubscript{ll}} & \textbf{OLS\textsubscript{l}} \\
\midrule
0.01 & 30.4 & 36.6 & 22.4 & 38.1 \\
0.10 & 32.8 & 37.5 & 23.9 & 39.7 \\
0.20 & 32.9 & 37.6 & 24.2 & 39.9 \\
0.30 & \underline{33.0} & 37.7 & 24.3 & 40.0 \\
0.40 & \underline{33.0} & 37.7 & 24.5 & 40.1 \\
0.50 & \textbf{33.1} & \underline{37.8} & 24.6 & \underline{40.2} \\
0.60 & \textbf{33.1} & \underline{37.8} & 24.8 & \underline{40.2} \\
0.70 & \textbf{33.1} & \underline{37.8} & \underline{24.9} & \textbf{40.3} \\
0.80 & \textbf{33.1} & \underline{37.8} & \textbf{25.0} & \textbf{40.3} \\
0.90 & \textbf{33.1} & \textbf{37.9} & \textbf{25.0} & \textbf{40.3} \\
0.95 & \textbf{33.1} & \textbf{37.9} & \textbf{25.0} & \textbf{40.3} \\
\bottomrule
\end{tabular}
}
\end{table}

To analyze the effect of thresholding on the mask probability map $\mathbf{M}_{\text{prob}}^{(l)}$, a range of values was evaluated to select grid points corresponding to each instance query. This threshold $\tau$~(Eq. \ref{eq: rasterization}) determines which points are retained from the gridized mask map. Table~\ref{table:threshold_analysis} presents the performance across DET\textsubscript{l}, DET\textsubscript{l\_ch}, TOP\textsubscript{ll}, and OLS\textsubscript{l} metrics for varying threshold values.

As shown in Table~\ref{table:threshold_analysis}, increasing the threshold improves performance across all metrics, with convergence observed around $\tau = 0.95$. This indicates that higher confidence regions in the mask probability map yield more reliable point selections for instance-level queries, enhancing centerline detection quality.

\subsection{Comparison of Lane Divider and Centerline Representations}
\label{sup_sec: lane_divider_vs_centerline}

\begin{table}[ht]
\caption{Chamfer distance-based mAP scores (DET\textsubscript{l\_ch}) of TopoBDA method for Centerline and Lane Divider representations across five splits in the Argoverse2 HDMap dataset. Best scores per split are highlighted in bold.}
\label{table: centerline_vs_lane_divider_different_splits}
\centering
\begin{tabular}{lccccc}
\toprule
\textbf{Representation} & \textbf{orig} & \textbf{near} & \textbf{farA} & \textbf{farB} & \textbf{farC} \\
\midrule
Centerline     & \textbf{39.8} & 25.0          & 20.3          & 19.1          & 25.3 \\
Lane Divider   & 38.9          & \textbf{27.6} & \textbf{22.5} & \textbf{20.8} & \textbf{25.5} \\
\bottomrule
\end{tabular}
\end{table}

In addition to centerline representation, lane dividers are also extracted from the HDMap to extend the scope of the analysis. Unlike the OpenLane-V2 dataset, which treats lane dividers as part of the lane segment concept, this work considers lane dividers as independent geometric entities as in studies \cite{liao_maptr_2023, li_hdmapnet_2022, liu_vectormapnet_2023}. Since HDMaps are structured around lane segments, converting them into lane dividers introduces duplication, particularly between adjacent lane segments. To address this, a chamfer distance-based elimination strategy is applied to remove redundant lane divider instances and ensure non-duplicate divider instances.

We conduct a comparative analysis of lane dividers and centerline representations across five distinct geographic splits of the Argoverse2 HDMap dataset. In this analysis, the Bezier head is utilized. The evaluation metric is the Chamfer distance-based mean Average Precision (DET\textsubscript{l\_ch}), and the results are summarized in Table~\ref{table: centerline_vs_lane_divider_different_splits}.

In the \textbf{Original} split, which exhibits significant geographic overlap between training and evaluation scenes, the centerline representation achieves the highest performance with an mAP of 39.8, slightly outperforming lane dividers at 38.9. However, in all other splits—\textbf{Near}, \textbf{FarA}, \textbf{FarB}, and \textbf{FarC}—lane dividers consistently outperform centerlines. For instance, in the \textbf{Near} split, lane dividers achieve an mAP of \textbf{27.6} compared to 25.0 for centerlines, and in the \textbf{FarA} split, lane dividers score \textbf{22.5} versus 20.3 for centerlines.

These results suggest that while centerline representations may benefit from memorization in geographically overlapping regions, they exhibit reduced generalization in unseen areas. In contrast, lane divider representations demonstrate more robust performance across diverse geographic domains, indicating stronger generalization capabilities and reduced susceptibility to overfitting.

\subsection{Comparison with SOTA on the Standard (Geographically Overlapping) Original Split}
\label{sup_sec: sota_original_split}

The \textbf{Original} split of OpenLane-V2 contains geographic overlap between training and
validation sets~\cite{yuan_streammapnet_2024, lilja_localization_2024}, which can inflate
reported performance through memorization (see Sec.~\ref{sec: comparison_sota} and
Sec.~\ref{sec: output_types_dataset_splits} in the main paper). We report results on this
standard benchmark for completeness and comparability with prior literature.
As discussed in the main paper, \textbf{TopoMaskV3~(F)} achieves 50.1 OLS --- the
second-highest score --- surpassing all methods except the Bezier-based
TopoBDA~\cite{kalfaoglu_topobda_2024}, whose advantage is attributable to
Bezier's higher susceptibility to geographic memorization.
The \textbf{TopoMaskV3~(M)} standalone head achieves 49.1 OLS, substantially outperforming
TopoMaskV2~(M) (46.3 OLS), validating the newly introduced offset and height prediction
mechanisms. This improvement surpasses most published works, including
TopoFormer~\cite{lv_t2sg_2024} and RoadPainter~\cite{ma_roadpainter_2024}.

\begin{table}[!t]
\centering
\caption{Comparative Evaluation of TopoMaskV3 and other Camera-Only Methods on the
\textbf{Original} split (\textbf{geographically overlapping}) of OpenLane-V2 Subset-A
using V1.1 Metric Baseline.}
\label{tab: sota_camera_only_subsetA}
\scalebox{0.7}{
\begin{tabular}{lccccc}
\toprule
\textbf{Method} & \textbf{DET\textsubscript{l}} & \textbf{DET\textsubscript{t}} &
\textbf{TOP\textsubscript{ll}} & \textbf{TOP\textsubscript{lt}} & \textbf{OLS} \\
\midrule
STSU \cite{can_structured_2021} & 12.7 & 43.0 & 2.9 & 19.8 & 29.3 \\
VectorMapNet \cite{liu_vectormapnet_2023} & 11.1 & 41.7 & 2.7 & 9.2 & 24.9 \\
MapTR \cite{liao_maptr_2023} & 8.3 & 43.5 & 2.3 & 8.3 & 24.2 \\
TopoNet \cite{li_graph-based_2023} & 28.6 & 48.6 & 10.9 & 23.9 & 39.8 \\
TopoMLP \cite{wu_topomlp_2024} & 28.5 & 49.5 & 21.7 & 26.9 & 44.1 \\
Topo2D \cite{li_enhancing_2024} & 29.1 & 50.6 & 22.3 & 26.2 & 44.4 \\
TopoLogic \cite{fu_topologic_2024} & 29.9 & 47.2 & 23.9 & 25.4 & 44.1 \\
RoadPainter \cite{ma_roadpainter_2024} & 30.7 & 47.7 & 22.8 & 27.2 & 44.6 \\
TopoFormer \cite{lv_t2sg_2024} & 34.7 & 48.2 & 24.1 & 29.5 & 46.3 \\
TopoMaskV2 (M) \cite{kalfaoglu_topomaskv2_2024} & 29.6 & \underline{53.8} & 20.6 & 31.9 & 46.3 \\
TopoMaskV2 (F) \cite{kalfaoglu_topomaskv2_2024} & 34.5 & \underline{53.8} & 24.5 & 35.6 & 49.4 \\
TopoBDA \cite{kalfaoglu_topobda_2024} & \textbf{38.9} & \textbf{54.3} & \textbf{27.6} & \textbf{37.3} & \textbf{51.7} \\
\textbf{TopoMaskV3 (M) (Ours)} & 34.7 & 53.4 & 23.8 & 35.4 & 49.1 \\
\textbf{TopoMaskV3 (F) (Ours)} & \underline{35.5} & 53.4 & \underline{25.9} & \underline{36.7} & \underline{50.1} \\
\bottomrule
\end{tabular}
}
\\ \footnotesize {(M): Mask-based, (F): Fusion-based approaches.}
\\ \footnotesize {Formatting: \textbf{bold} = best, \underline{underline} = second-best.}
\end{table}